\documentclass[journal,comsoc]{IEEEtran}
\usepackage[T1]{fontenc}% optional T1 font encoding

\ifCLASSINFOpdf
   \usepackage[pdftex]{graphicx}
\else
\fi
\usepackage{amsmath}
\usepackage[cmintegrals]{newtxmath}
\newtheorem{theorem}{Theorem}
\newtheorem{definition}{Definition}

\newtheorem{proposition}{Proposition}
\newtheorem{remark}{Remark}
\usepackage{bm}
\usepackage{algorithm}  
\usepackage{algorithmicx}  
\usepackage{algpseudocode}  
\usepackage{amsmath}  

\usepackage{float}  
\usepackage{lipsum}
\makeatletter
\usepackage{array}
\usepackage{multirow}
\usepackage{longtable}
\usepackage{rotating}
\usepackage{stfloats}
\usepackage{diagbox}
\usepackage{graphicx}  %插入图片的宏包
\ifCLASSOPTIONcompsoc
\usepackage[caption=false,font=normalsize,labelfon
t=sf,textfont=sf]{subfig}
\else
\usepackage[caption=false,font=footnotesize]{subfig}
\fi

\begin{document}
\title{Low-Rank Tensor Completion Based on Bivariate Equivalent Minimax-Concave Penalty}

\author{Hongbing Zhang,
        Xinyi Liu, Hongtao Fan, Yajing Li, Yinlin Ye% <-this % stops a space
%\thanks{This work was supported in part by}% <-this % stops a space
% <-this % stops a space
\thanks{This work was supported by the National Natural Science Foundation of
	China (Nos. 11701456, 11801452, 11571004), Fundamental Research Project of Natural Science in Shaanxi Province General Project (Youth) (Nos. 2019JQ-415, 2019JQ-196), the Initial Foundation for Scientific Research of Northwest A\&F University (Nos.2452017219, 2452018017), and Innovation and Entrepreneurship Training Program for College Students of Shaanxi Province (S201910712132).}
%\thanks{This work was supported, in partial, by Natural Science Basic Research Program of Shaanxi (Program Nos. XXXXXXX,
%	XXXXXXX, XXXXXX), and the Initial Foundation for Scientific Research
%	of Northwest A\&F University (Program Nos. XXXXXXX, XXXXXXX).}
\thanks{H. Zhang, X. Liu, H. Fan, Y. Li and Y. Ye are with Department of information and Computing Science, College of Science, Northwest A\&F University, Yangling, Shaanxi 712100, China(e-mail: zhanghb@nwafu.edu.cn; Lxy6x1@163.com; fanht17@nwafu.edu.cn; hliyajing@163.com; 13314910376@163.com)}.
}

\markboth{}%
{Shell \MakeLowercase{\textit{et al.}}: Bare Demo of IEEEtran.cls for IEEE Communications Society Journals}
\maketitle 

\begin{abstract}
Low-rank tensor completion (LRTC) is an important problem in computer vision and machine learning. The minimax-concave penalty (MCP) function as a non-convex relaxation has achieved good results in the LRTC problem. To makes all the constant parameters of the MCP function as variables so that futherly improving the adaptability to the change of singular values in the LRTC problem, we propose the bivariate equivalent minimax-concave penalty (BEMCP) theorem. Applying the BEMCP theorem to tensor singular values leads to the bivariate equivalent weighted tensor $\Gamma$-norm (BEWTGN) theorem, and we analyze and discuss its corresponding properties. Besides, to facilitate the solution of the LRTC problem, we give the proximal operators of the BEMCP theorem and BEWTGN. Meanwhile, we propose a BEMCP model for the LRTC problem, which is optimally solved based on alternating direction multiplier (ADMM). Finally, the proposed method is applied to the data restorations of multispectral image (MSI), magnetic resonance imaging (MRI) and color video (CV) in real-world, and the experimental results demonstrate that it outperforms the state-of-arts methods.
\end{abstract}


\IEEEpeerreviewmaketitle

\section{INTRODUCTION}

\IEEEPARstart{W}{ith} the rapid development of information technology, researchers increasingly encounter real data with high dimensions and complex structures. Tensors, as high-dimensional generalizations of vectors and matrices, can better represent the complex properties of high-dimensional data and play an increasingly important role in many applications, such as color image/video (CI/CV) processing \cite{1472018163}, \cite{3762018t397}, \cite{7562018767}, \cite{1372020149}, hyperspectral/multispectral image (HSI/MSI) processing \cite{8359412}, \cite{7467446}, \cite{8657368}, \cite{8941238}, magnetic resonance imaging (MRI) data recovery \cite{1242020783}, \cite{4032018417}, \cite{9412019964}, background subtraction \cite{5032019109}, \cite{8319458}, \cite{7488247}, video rain stripe removal \cite{8237537}, \cite{8578793} and signal reconstruction \cite{7010937}, \cite{7676397}.

Low-rank tensor completion (LRTC) is an important issue in tensor recovery research. The general method expresses the low-rank tensor completion problem as follows:
\begin{eqnarray}
\min_{\mathcal{X}}rank(\mathcal{X})\quad s.t.\quad  \mathcal{P}_{\Omega}(\mathcal{X}-\mathcal{Y})=\mathbf{0},
\label{lossfunction}\end{eqnarray}
where $\mathcal{Y}\in\mathbb{R}^{\mathit{I}_{1}\times\mathit{I}_{2}\times\cdots\times\mathit{I}_{N}}$ is the observation, $\Omega$ is the index set for the known entries, and $\mathcal{P}_{\Omega}(\mathcal{Y})$ is a projection
operator that keeps the entries of $\mathcal{Y}$ in $\Omega$ and sets all others to zero, $rank(\mathcal{X})$ defines the tensor rank of $\mathcal{X}$. In fact, tensors differ from matrices in that the definition of their rank is not unique. In the past decades, the most popular definitions of rank are CANDECOMP/PARAFAC(CP) rank based on CP decomposition \cite{41201156}, \cite{36232017} and Tucker rank based on Tucker decomposition \cite{64201881}, \cite{7460200}, as well as tubal rank and multi-rank based on t-SVD \cite{6909886}. Solving the CP rank problem of tensors is a NP-hard \cite{139201360}, which is not conducive to better application. The calculation of Tucker rank requires data to be folded and unfolded, which will cause structural damage to data. Compared with CP rank and Tucker rank, the tubal rank and multiple rank obtained based on t-SVD can better maintain the data structure, but its tensor-tensor product limitation prevents it from being applied to higher order cases. Recently, Zheng et al. \cite{2020170} proposed a new form of rank (N-tubal rank) based on tubal rank, which adopts a new unfold method of higher-order tensors into third-order tensors in various directions. This approach makes good use of the properties of tensor tubal rank but also enables t-SVD to be applied to higher-order cases. Therefore, because of the excellent properties of N-tubal rank, we will also consider N-tubal rank to construct the model we propose in this paper.

Undoubtedly, the development of computationally efficient algorithms to solve problem (\ref{lossfunction}) is of great practical value. However, the rank optimization problem in problem (\ref{lossfunction}) will lead to NP-hard problem \cite{139201360}, which will seriously affect the efficiency of solving the problem. In this regard, researchers turn to its convex relaxation or non-convex relaxation forms. For convex relaxation, although it is easier to solve, it will produce biased estimates \cite{448120174494}. It is relatively difficult to solve for non-convex relaxations, but leads to more accurate results \cite{2271995234}, \cite{211720122130}, \cite{4712010501}. Recently, the minimax-concave penalty function as a non-convex relaxation has achieved good results in the LRTC problem \cite{qiu2021nonlocal}, \cite{2920151}. Further research on this function has profound implications for the LRTC problem, and its definition is as follows:
\begin{definition}[Minimax-concave penalty (MCP) function \cite{8942010942}]
	Let $\lambda>0,\gamma>1$. The minimax-concave penalty function $h_{\gamma,\lambda}:\mathbb{R}\to\mathbb{R}_{\geqslant0}$ is defined as 
	\begin{eqnarray}
	h_{\gamma,\lambda}(y)=\left\{\begin{array}{c}
	\lambda\rvert y\rvert-\frac{y^{2}}{2\gamma},\qquad \rvert y\rvert\leqslant\gamma\lambda,
	\\\frac{\lambda^{2}}{2\gamma},\qquad\qquad\quad \rvert y\rvert\geqslant\gamma\lambda.
	\end{array}
	\right.\label{scalarMCP}
	\end{eqnarray}
\end{definition}

It is not difficult to find that the MCP function contains two constant parameters, i.e., $\lambda$ and $\gamma$. It is considered that in the LRTC problem when the MCP function acts on the tensor singular values, its parameters are fixed. But the tensor singular values will change with the iteration update. Recently, an equivalent MCP (EMCP) method was proposed in \cite{2132021245}, which is a novel non-convex relaxation method based on the traditional MCP method. The EMCP transforms the parameter $\lambda$ in MCP into a variable form through the equivalence theorem so that it can adapt to the change of tensor singular values. It is wroth noting that the MCP function contains two parameters, i.e., $\lambda$ and $\gamma$, which, in reality, affect each other, and the nonconvex relaxations produced by different $\lambda$ and $\gamma$ vary widely. A key idea is that it is especially important to expect to turn both parameters into variables at the same time, resulting in a more efficient equivalence theorem. Motivated by this, we propose a new structural equivalence theorem, i.e., the bivariate equivalent minimax-concave penalty (BEMCP) theorem, that allows $\lambda$ and $\gamma$ to be transformed into variables at the same time. The difference between MCP, EMCP, and BEMCP can be seen from the Table \ref{3CASE}.
\begin{table}[h]
	\caption{Variable case for three methods}
\centering
	\large
	\begin{tabular}{|c|c|c|}
		
		\hline
		\diagbox{Method}{Variable}& $\lambda$ & $\gamma$ \\ \hline
		MCP   & $\times$      & $\times$     \\ \hline
		EMCP  & {\small $\surd$}     & $\times$     \\ \hline
		BEMCP &  {\small $\surd$}      & {\small $\surd$}     \\ \hline
\end{tabular}
\\
{\large \text{The symbols
		 "{\small $\surd$}" and "$\times$" indicate whether}}
	 \\\text{  $\lambda$ and $\gamma$ are variables or not.}
\label{3CASE}
\end{table}

To sum up, the main contributions of our paper are:

Firstly, a new structural equivalence theorem called BEMCP theorem is proposed, which turns two constant parameters $\lambda$ and $\gamma$ into variables at the same time and further improves adaptability of tensor singular value change in the LRTC problem. Applying the BEMCP theorem to tensor singular values leads to the bivariate equivalent weighted tensor $\Gamma$-norm (BEWTGN) theorem, and the corresponding properties are analyzed and discussed. Furthermore, to solve the established model based on this new theorem, the corresponding proximal operators of the BEMCP theorem and BEWTGN are proposed.  

Secondly, for the LRTC problem, we propose a new model, i.e., BEMCP model based on N-tubal rank. Furthermore, we design an efficient alternating direction multiplier method (ADMM) algorithm \cite{683201156}, \cite{6122011620} to optimally solve these problems. On this basis, the closed solution of each variable update is deduced, so that the algorithm can be executed efficiently.

Thirdly, three different types of data, i.e., MSI, MRI, and CV, are used to verify the effectiveness and efficiency of proposed method. Extensive numerical experiments demonstrate that the results obtained by our method have clear advantages over the comparative method in both visual and quantitative values. 

The summary of this article is as follows: In Section II, some preliminary knowledge and background of the tensors are given. The theorems about BEMCP and its properties are presented in Section III. In Section IV, we give the corresponding proximal operators and proofs of the BEMCP theorem and BEWTGN. The main results, including the proposed model and algorithm, are shown in Section V. The results of extensive experiments and discussions are presented in Section VI. Conclusions are drawn in Section VII.

\section{PRELIMINARIES}
%This section provides the basic ingredients to induce the proposed method. 
%This section provides the basic knowledge of the proposed method. Firstly, we give the basic tensor notations.
\subsection{Tensor Notations and Definitions}
In this section, we give some basic notations and briefly introduce some definitions used throughout the paper. Generally, a lowercase letter and an uppercase letter denote a vector $y$ and a marix $Y$, respectively. An $N$th-order tensor is denoted by a calligraphic uppercase letter $\mathcal{Y}\in \mathbb{R}^{\mathit{I}_{1}\times\mathit{I}_{2}\times\cdots\times\mathit{I}_{N}}$ and $\mathcal{Y}_{i_{1},i_{2},\cdots,i_{N}}$ is its $(i_{1},i_{2},\cdots,i_{N})$-th element. The Frobenius norm of a tensor is defined as $\|\mathcal{Y}\|_{F}=(\sum_{i_{1},i_{2},\cdots,i_{N}}\mathit{y}_{i_{1},i_{2},\cdots,i_{N}}^{2})^{1/2}$. For a three order tensor $\mathcal{Y}\in\mathbb{R}^{\mathit{I}_{1}\times\mathit{I}_{2}\times\mathit{I}_{3}}$, we use $\bar{\mathcal{Y}}$ to denote along each tubal of $\mathcal{Y}$, i.e., $\bar{\mathcal{Y}}=fft(\mathcal{Y},[],3)$. The inverse DFT is computed by command $ifft$ satisfying $\mathcal{Y}=ifft(\bar{\mathcal{Y}},[],3)$. More often, the frontal slice $\mathcal{Y}(:,:,i)$ is denoted compactly as $\mathcal{Y}^{(i)}$. The Hadamard product is the elementwise tensor product. Given tensors $\mathcal{A}$ and $\mathcal{B}$, both of size $\mathit{I}_{1}\times\mathit{I}_{2}\times\cdots\times\mathit{I}_{N}$, their Hadamard product is denoted by $\mathcal{A}\star \mathcal{B}$. And the elementwise tensor division is denoted by $\frac{\mathcal{A}}{\mathcal{B}}$.  The set of real numbers greater than $b$ real numbers is denoted as $\mathbb{R}_{\geqslant b}=\{\upsilon\in\mathbb{R}\quad\vert\quad\upsilon\geqslant b\}$. The set of tensors consisting of all real-valued elements greater than $b$ can be expressed as $\mathbb{R}^{\mathit{I}_{1}\times\mathit{I}_{2}\times\cdots\times\mathit{I}_{N}}_{\geqslant b}=\{\bar{\upsilon}\in\mathbb{R}^{\mathit{I}_{1}\times\mathit{I}_{2}\times\cdots\times\mathit{I}_{N}}\quad\vert\quad\bar{\upsilon}_{i_{1},i_{2},\cdots,i_{N}}\geqslant b\}$.

\begin{definition}[Mode-$k_{1}k_{2}$ slices \cite{2020170}]
	For an $N$th-order tensor $\mathcal{Y}\in \mathbb{R}^{\mathit{I}_{1}\times\mathit{I}_{2}\times\cdots\times\mathit{I}_{N}}$, its mode-$k_{1}k_{2}$ slices ($\mathcal{Y}^{(k_{1}k_{2})},1\leqslant k_{1} <k_{2}\leqslant N,k_{1},k_{2}\in\mathbb{Z}$) are two-dimensional sections, defined by fixing all but the mode-$k_{1}$ and the mode-$k_{2}$ indexes.
\end{definition}
\begin{definition}[Tensor Mode-$k_{1},k_{2}$ Unfolding and Folding \cite{2020170}]
	For an $N$th-order tensor $\mathcal{Y}\in \mathbb{R}^{\mathit{I}_{1}\times\mathit{I}_{2}\times\cdots\times\mathit{I}_{N}}$, its mode-$k_{1}k_{2}$ unfolding is a three order tensor denoted by $\mathcal{Y}_{(k_{1}k_{2})}\in\mathbb{R}^{\mathit{I}_{k_{1}}\times\mathit{I}_{k_{2}}\times\prod_{s\neq k_{1},k_{2}}\mathit{I}_{s}}$, the frontal slices of which are the lexicographic orderings of the mode-$k_{1}k_{2}$ slices of $\mathcal{Y}$. Mathematically, the  $(i_{1}, i_{2},...,i_{N} )$-th element of $\mathcal{Y}$ maps to the $(i_{k_{1}},i_{k_{2}},j)$-th element of $\mathcal{Y}_{(k_{1}k_{2})}$, where
	\begin{equation}
	j=1+\sum_{s=1,s\neq k_{1},s\neq k_{2}}^{N}(i_{s}-1)\mathit{J}_{s}\quad with\quad \mathit{J}_{s}=\prod_{m=1,m\neq k_{1},m\neq k_{2}}^{s-1}\mathit{I}_{m}. 
	\end{equation}
	The mode-$k_{1}k_{2}$ unfolding operator and its inverse operation are respectively represented as $\mathcal{Y}_{(k_{1}k_{2})}:=t-unfold(\mathcal{Y},k_{1},k_{2})$ and $\mathcal{Y}:=t-fold(\mathcal{Y}_{(k_{1}k_{2})},k_{1},k_{2})$.
\end{definition}

For a three order tensor $\mathcal{Y}\in\mathbb{R}^{\mathit{I}_{1}\times\mathit{I}_{2}\times\mathit{I}_{3}}$, the block circulation operation is defined as
\begin{equation}
bcirc(\mathcal{Y}):=
\begin{pmatrix}
\mathcal{Y}^{(1)}& \mathcal{Y}^{(\mathit{I}_{3})}&\dots& \mathcal{Y}^{(2)}&\\
\mathcal{Y}^{(2)}& \mathcal{Y}^{(1)}&\dots& \mathcal{Y}^{(3)}&\\
\vdots&\vdots&\ddots&\vdots&\\
\mathcal{Y}^{(\mathit{I}_{3})}& \mathcal{Y}^{(\mathit{I}_{3}-1)}&\dots& \mathcal{Y}^{(1)}&
\end{pmatrix}\in\mathbb{R}^{\mathit{I}_{1}\mathit{I}_{3}\times\mathit{I}_{2}\mathit{I}_{3}}.\nonumber
\end{equation}

The block diagonalization operation and its inverse operation are given by 
\begin{eqnarray}
&&bdiag(\mathcal{Y}):=\begin{pmatrix}
\mathcal{Y}^{(1)} & & &\\
& \mathcal{Y}^{(2)} & &\\
& & \ddots &\\
& & & \mathcal{Y}^{(\mathit{I}_{3})}
\end{pmatrix} \in\mathbb{R}^{\mathit{I}_{1}\mathit{I}_{3}\times\mathit{I}_{2}\mathit{I}_{3}},\nonumber\\
&&bdfold(bdiag(\mathcal{Y})):=\mathcal{Y}.\nonumber
\end{eqnarray}

The block vectorization operation and its inverse operation are defined as 
\begin{eqnarray}
bvec(\mathcal{Y}):=\begin{pmatrix}
\mathcal{Y}^{(1)}\\\mathcal{Y}^{(2)}\\\vdots\\\mathcal{Y}^{(\mathit{I}_{3})}
\end{pmatrix}\in\mathbb{R}^{\mathit{I}_{1}\mathit{I}_{3}\times\mathit{I}_{2}},\quad bvfold(bvec(\mathcal{Y})):=\mathcal{Y}.\nonumber
\end{eqnarray}
\begin{definition}[T-product \cite{6416568}]
	Let $\mathcal{A}\in\mathbb{R}^{\mathit{I}_{1}\times\mathit{I}_{2}\times\mathit{I}_{3}}$ and $\mathcal{B}\in\mathbb{R}^{\mathit{I}_{2}\times\mathit{J}\times\mathit{I}_{3}}$. Then the t-product $\mathcal{A}\ast\mathcal{B}$ is defined to be a tensor of size $\mathit{I}_{1}\times\mathit{J}\times\mathit{I}_{3}$,
	\begin{eqnarray}
	\mathcal{A}\ast\mathcal{B}:=bvfold(bcirc(\mathcal{A})bvec(\mathcal{B})).\nonumber
	\end{eqnarray}
	
	Since that circular convolution in the spatial domain is equivalent to multiplication in the Fourier domain, the T-product between two tensors $\mathcal{C}=\mathcal{A}\ast\mathcal{B}$ is equivalent to
	\begin{eqnarray}
	\bar{\mathcal{C}}=bdfold(bdiag(\bar{\mathcal{A}})bdiag(\bar{\mathcal{B}})).\nonumber
	\end{eqnarray}
\end{definition}
%\begin{figure}
%	\centering
%	\includegraphics[width=0.7\linewidth]{images/TSVD1}
%	\caption{An illustration of the t-SVD of an $\mathit{I}_{1}\times\mathit{I}_{2}\times\mathit{I}_{3}$ tensor}
%	\label{tsvd}
%\end{figure}
\begin{definition}[Tensor conjugate transpose \cite{6416568}]
	The conjugate transpose of a tensor $\mathcal{A}\in\mathbb{C}^{\mathit{I}_{1}\times\mathit{I}_{2}\times\mathit{I}_{3}}$ is the tensor $\mathcal{A}^{H}\in\mathbb{C}^{\mathit{I}_{2}\times\mathit{I}_{1}\times\mathit{I}_{3}}$ obtained by conjugate transposing each of the frontal slices and then reversing the order of transposed frontal slices 2 through $\mathit{I}_{3}$.
\end{definition}
\begin{definition}[Identity tensor \cite{6416568}]
	The identity tensor $\mathcal{I}\in\mathbb{R}^{\mathit{I}_{1}\times\mathit{I}_{1}\times\mathit{I}_{3}}$ is the tensor whose first frontal slice is the $\mathit{I}_{1}\times\mathit{I}_{1}$ identity matrix, and whose other frontal slices are all zeros.
\end{definition}

It is clear that $bcirc(\mathcal{I})$ is the  $\mathit{I}_{1}\mathit{I}_{3}\times\mathit{I}_{1}\mathit{I}_{3}$ identity matrix. So it is easy to get $\mathcal{A}\ast\mathcal{I}=\mathcal{A}$ and $\mathcal{I}\ast\mathcal{A}=\mathcal{A}$. 
\begin{definition}[Orthogonal tensor \cite{6416568}]
	A tensor $\mathcal{Q}\in\mathbb{R}^{\mathit{I}_{1}\times\mathit{I}_{1}\times\mathit{I}_{3}}$ is orthogonal if it satisfies
	\begin{eqnarray}
	\mathcal{Q}\ast\mathcal{Q}^{H}=\mathcal{Q}^{H}\ast\mathcal{Q}=\mathcal{I}.\nonumber
	\end{eqnarray}
\end{definition}
\begin{definition}[F-diagonal tensor \cite{6416568}]
	A tensor is called f-diagonal if each of its frontal slices is a diagonal matrix.
\end{definition}
\begin{theorem}[t-SVD \cite{8606166}]
	Let $\mathcal{X}\in\mathbb{R}^{\mathit{I}_{1}\times\mathit{I}_{2}\times\mathit{I}_{3}}$ be a three order tensor, then it can be factored as 
	\begin{eqnarray}
	\mathcal{X}=\mathcal{U}\ast\mathcal{S}\ast\mathcal{V}^{H},\nonumber
	\end{eqnarray}
	where $\mathcal{U}\in\mathbb{R}^{\mathit{I}_{1}\times\mathit{I}_{1}\times\mathit{I}_{3}}$ and $\mathcal{V}\in\mathbb{R}^{\mathit{I}_{2}\times\mathit{I}_{2}\times\mathit{I}_{3}}$ are orthogonal tensors, and $\mathcal{S}\in\mathbb{R}^{\mathit{I}_{1}\times\mathit{I}_{2}\times\mathit{I}_{3}}$ is an f-diagonal tensor. 
%	The t-SVD scheme is illustrated in Fig.\ref{tsvd}, and its computation is given in Algorithm \ref{TSVD1}.
\end{theorem}
%\begin{algorithm}[t]
%	\caption{t-SVD \cite{8606166}} %算法的名字
%	\hspace*{0.02in} {\bf Input:} %算法的输入,  \hspace*{0.02in}用来控制位置, 同时利用 \\ 进行换行
%	$\mathcal{Y}\in\mathbb{R}^{\mathit{I}_{1}\times\mathit{I}_{2}\times\mathit{I}_{3}}$ \\
%	\hspace*{0.02in} {\bf Output:} %算法的结果输出
%	t-SVD components $\mathcal{U}$, $\mathcal{S}$ and $\mathcal{V}$ of $\mathcal{A}$.
%	\begin{algorithmic}
%		\State Compute $\bar{\mathcal{A}}=fft(\mathcal{A},[],3)$.
%		%\State some description % \State 后写一般语句
%		\State Compute each frontal slice of $\bar{\mathcal{U}}$, $\bar{\mathcal{S}}$, and $\bar{\mathcal{V}}$ from $\bar{\mathcal{A}}$ by
%		\For {$i=1,\dots,[\frac{\mathit{I}_{3}+1}{2}]$} 
%		\State $[\bar{\mathcal{U}}^{(i)}, \bar{\mathcal{S}}^{(i)}, \bar{\mathcal{V}}^{(i)}]=SVD(\bar{\mathcal{A}}^{(i)})$;
%		\EndFor
%		\For {$i=[\frac{\mathit{I}_{3}+1}{2}]+1,\dots,\mathit{I}_{3}$} 
%		\State $[\bar{\mathcal{U}}^{(i)}, \bar{\mathcal{S}}^{(i)}, \bar{\mathcal{V}}^{(i)}]=SVD(\bar{\mathcal{A}}^{(i)})$;
%		\State $\bar{\mathcal{U}}^{(i)}=conj(\bar{\mathcal{U}}^{(\mathit{I}_{3}-i+2)})$;
%		\State $\bar{\mathcal{S}}^{(i)}=\bar{\mathcal{S}}^{(\mathit{I}_{3}-i+2)}$;
%		\State $\bar{\mathcal{V}}^{(i)}=conj(\bar{\mathcal{V}}^{(\mathit{I}_{3}-i+2)})$;
%		\EndFor
%	\end{algorithmic}
%	\hspace*{0.02in}  %算法的结果输出
%	Compute $\mathcal{U}=ifft(\bar{\mathcal{U}},[],3)$, $\mathcal{S}=ifft(\bar{\mathcal{S}},[],3)$, and $\mathcal{V}=ifft(\bar{\mathcal{V}},[],3)$. 
%	\label{TSVD1}\end{algorithm}
\begin{definition}[Tensor tubal-rank and multi-rank \cite{6909886}]
	The tubal-rank of a tensor $\mathcal{Y}\in\mathbb{R}^{\mathit{I}_{1}\times\mathit{I}_{2}\times\mathit{I}_{3}}$, denoted as $rank_{t}(\mathcal{Y})$, is defined to be the number of non-zero singular tubes of $\mathcal{S}$, where $\mathcal{S}$ comes from the t-SVD of $\mathcal{Y}:\mathcal{Y}=\mathcal{U}\ast\mathcal{S}\ast\mathcal{V}^{H}$. That is 
	\begin{eqnarray}
	rank_{t}(\mathcal{Y})=\#\{i:\mathcal{S}(i,:,:)\neq0\}.
	\end{eqnarray}
	The tensor multi-rank of $\mathcal{Y}\in\mathbb{R}^{\mathit{I}_{1}\times\mathit{I}_{2}\times\mathit{I}_{3}}$ is a vector, denoted as $rank_{r}(\mathcal{Y})\in\mathbb{R}^{\mathit{I}_{3}}$, with the $i$-th element equals to the rank of $i$-th frontal slice of $\mathcal{Y}$.
\end{definition}

\begin{definition}[Tensor nuclear norm (TNN)]
	The tensor nuclear norm of a tensor $\mathcal{Y}\in\mathbb{R}^{\mathit{I}_{1}\times\mathit{I}_{2}\times\mathit{I}_{3}}$, denoted as $\|\mathcal{Y}\|_{TNN}$, is defined as the sum of the singular values of all the frontal slices of $\bar{\mathcal{Y}}$, i.e.,
	\begin{eqnarray}
	\|\mathcal{Y}\|_{TNN}:=\sum_{i=1}^{\mathit{I}_{3}}\|\bar{\mathcal{Y}}^{(i)}\|_{\ast}
	\end{eqnarray}
	where $\bar{\mathcal{Y}}^{(i)}$ is the $i$-th frontal slice of $\bar{\mathcal{Y}}$, with $\bar{\mathcal{Y}}=fft(\mathcal{Y},[],3)$.
\end{definition}

\begin{definition}[N-tubal rank \cite{2020170}]
	The N-tubal rank of an Nth-order tensor $\mathcal{Y}\in \mathbb{R}^{\mathit{I}_{1}\times\mathit{I}_{2}\times\cdots\times\mathit{I}_{N}}$ is defined as a vector, the elements of which
	contain the tubal rank of all mode-$k_{1}k_{2}$ unfolding tensors, i.e.,
    \begin{eqnarray}
	N-rank_{t}(\mathcal{Y}):=(rank_{t}(\mathcal{Y}_{(12)}),rank_{t}(\mathcal{Y}_{(13)}),\cdots,\nonumber\\rank_{t}(\mathcal{Y}_{(1N)}),rank_{t}(\mathcal{Y}_{(23)}),\cdots,\nonumber rank_{t}(\mathcal{Y}_{(2N)}),\cdots,\\rank_{t}(\mathcal{Y}_{(N-1N)}))\in\mathbb{R}^{N(N-1)/2}.
	\end{eqnarray}
\end{definition}
\begin{theorem}[Equivalent minimax-concave penalty (EMCP) \cite{2132021245}]
	Let $\lambda\in\mathbb{R}_{>0},\gamma>1$ and $y\in\mathbb{R}$. The MCP $h_{\gamma,\lambda}:\mathbb{R}\to\mathbb{R}_{\geqslant0}$ is the solution of the following optimization problem:
	\begin{eqnarray}
	h_{\gamma,\lambda}(y)=\min_{\omega\in\mathbb{R}_{\geqslant0}} \left\lbrace \omega\rvert y\rvert+\frac{\gamma}{2}(\omega-\lambda)^{2}\right\rbrace .
	\end{eqnarray}
\end{theorem}
\section{BIVARIATE EQUIVALENT MINIMAX-CONCAVE PENALTY}
In this section, we will construct and obtain the BEMCP theorem, which turns $\lambda$ and $\gamma$ into variables at the same time. Then the BEWTGN theorem is applied to tensor singular values to deduce the BEWTGN theorem, and its corresponding properties are analyzed and established. 
%Applying the BEMCP theorem to tensor singular values leads to the BEWTGN theorem, and its corresponding properties are analyzed and discussed.
	\begin{theorem}[Bivariate Equivalent Minimax-Concave Penalty (BEMCP)]
	Let $\lambda, \upsilon\in\mathbb{R}_{\geqslant0}, \gamma\in\mathbb{R}_{>1}$ and $y\in\mathbb{R}$. The MCP $h_{\gamma,\lambda}:\mathbb{R}\to\mathbb{R}_{\geqslant0}$ is the solution of the following optimization problem:
	\begin{eqnarray}
	h_{\gamma,\lambda}(y)=\min_{\upsilon\geqslant0}\dfrac{2\upsilon\rvert y\rvert+(\upsilon-\lambda\gamma)^{2}}{2\gamma}.\label{BEMCP}
	\end{eqnarray}
\label{tbemcp}\end{theorem}	
\begin{IEEEproof}
	Consider the following function
	\begin{eqnarray}
	h(y,\upsilon)=\dfrac{2\upsilon\rvert y\rvert+(\upsilon-\lambda\gamma)^{2}}{2\gamma}.
	\end{eqnarray}
	Let $\upsilon^{\star}$ denote the first-order critical point of $h(y,\upsilon)$, i.e., 
	\begin{eqnarray}
	\upsilon^{\star}=\arg\min_{\upsilon}h(y,\upsilon).
	\end{eqnarray}
	Since $h(y,\upsilon)$ is differentiable with respect to $\upsilon$, setting
	\begin{eqnarray}
	\dfrac{\partial h(y,\upsilon)}{\partial\upsilon}\rvert_{\upsilon=\upsilon^{\star}}=0
	\end{eqnarray}
	gives 
	\begin{eqnarray}
	\upsilon^{\star}=\left\{\begin{array}{c}
	\lambda\gamma-\rvert y\rvert,\quad \rvert y\rvert\leqslant\gamma\lambda,
	\\0,\qquad\quad\,\, \rvert y\rvert\geqslant\gamma\lambda.
	\end{array}
	\right.
	\end{eqnarray}
	The BEMCP is given by $h_{\gamma,\lambda}(y)=h(y,\upsilon^{\star})$. Substituting for $\upsilon^{\star}$ in $h(y,\upsilon)$, we get 
	\begin{eqnarray}
	h_{\gamma,\lambda}(y)=\left\{\begin{array}{c}
	\lambda\rvert y\rvert-\frac{y^{2}}{2\gamma},\qquad\, \rvert y\rvert\leqslant\gamma\lambda,
	\\\frac{\lambda^{2}}{2\gamma},\qquad\qquad\quad \rvert y\rvert\geqslant\gamma\lambda.
	\end{array}
	\right.
	\end{eqnarray}
	
%	Therefore the BEMCP theorem is verified.
\end{IEEEproof}
%\begin{remark}
%	Unlike EMCP in \cite{2132021245}, we use $\upsilon$ to replace $\lambda\gamma$, making both $\lambda$ and $\gamma$ into variables. In EMCP, $\omega$ completely replaces $\lambda$, but we still keep $\gamma$ in our equivalent structure, and do not replace both $\lambda$ and $\gamma$ in the part related only to $\upsilon$.
%\end{remark}

\begin{definition}[Weighted Tensor $\Gamma$-norm (WTGN) ]
	The tensor $\Gamma$-norm of $\mathcal{Y}\in\mathbb{R}^{\mathit{I}_{1}\times\mathit{I}_{2}\times\mathit{I}_{3}}$, denoted by $\|\mathcal{Y}\|_{\Gamma,\Lambda}$, is defined as follows:
	\begin{eqnarray}
	\|\mathcal{Y}\|_{\Gamma,\Lambda}=\sum_{i=1}^{\mathit{I}_{3}}\sum_{j=1}^{R}h_{\Gamma_{(i,j)}, \Lambda_{(i,j)}}(\sigma_{j}(\bar{\mathcal{Y}}^{(i)})),  \label{wtgnmcp}
	\end{eqnarray}
	where $R=\min(\mathit{I}_{1},\mathit{I}_{2})$, $\Lambda\in\mathbb{R}^{\mathit{I}_{3}\times R}_{\geqslant0},\Gamma\in\mathbb{R}^{\mathit{I}_{3}\times R}_{>1}$ .
\end{definition}

\begin{theorem}[Bivariate Equivalent Weighted Tensor $\Gamma$-norm (BEWTGN)]
	For a third-order tensor $\mathcal{Y}\in\mathbb{R}^{\mathit{I}_{1}\times\mathit{I}_{2}\times\mathit{I}_{3}}$. Let $\upsilon,\bar{\Lambda}\in\mathbb{R}^{\mathit{I}_{3}\times R}_{\geqslant0},\Gamma\in\mathbb{R}^{\mathit{I}_{3}\times R}_{>1}$, and $R=\min\{\mathit{I}_{1},\mathit{I}_{2}\}$. The weighted tensor $\Gamma$-norm is obtained equivalently as 
	\begin{eqnarray}
	\|\mathcal{Y}\|_{\Gamma,\Lambda}=\min_{\upsilon} \left\lbrace \|\mathcal{Y}\|_{\frac{\upsilon}{\Gamma},\ast}+\frac{1}{2}\|\frac{\upsilon-\Lambda\star\Gamma}{\Gamma\star\Gamma}\|_{F}^{2}\right\rbrace \label{ewtgn}
	\end{eqnarray}          
where $\|\mathcal{Y}\|_{\frac{\upsilon}{\Gamma},\ast}=\sum_{i_{3}=1}^{\mathit{I}_{3}}\sum_{j=1}^{R}\frac{\upsilon_{(i_{3},j)}}{\Gamma_{(i_{3},j)}}\sigma_{j}(\bar{\mathcal{Y}}^{(i_{3})})$ is the weighted nuclear-norm of $i_{3}$-th slice of $\bar{\mathcal{Y}}$, and $\bar{\mathcal{Y}}=fft(\mathcal{Y}, [], 3)$.
\end{theorem}
\begin{IEEEproof}
	The proof is similar to that of Theorem \ref{tbemcp}, since
	the objective function in Eq.(\ref{ewtgn}) is non-negative and separable.
%	 in the entries of the weight matrix $\frac{\upsilon}{\Gamma}$.
\end{IEEEproof}
\begin{remark}
	In particular, when the third dimension $\mathit{I}_{3}$ of the third-order tensor $\mathcal{Y}$ is 1, the BEWTGN can degenerate into the form of the bivariate equivalent matrix $\Gamma$-norm.
\end{remark}

\begin{remark}
	Unlike the nuclear norm penalty, the WTGN(\ref{wtgnmcp}), and the BEWTGN (\ref{ewtgn}) do not satisfy the triangle inequality. Some important properties of the BEWTGN itself are presented below.
\end{remark}

\begin{proposition}
	The BEWTGN defined in (\ref{ewtgn}) satisfies the following properties:
	\\\textbf{(a) Non-negativity}: The BEWTGN is non-negative, i.e., $\|\mathcal{Y}\|_{\Gamma,\Lambda}\geqslant0$. The equality holds if and only if $\mathcal{Y}$ is the null tensor.
	\\\textbf{(b) Concavity}:
	$\|\mathcal{Y}\|_{\Gamma,\Lambda}$ is concave in the modulus of the singular values of $\mathcal{Y}$.
	\\\textbf{(c) Boundedness}: The BEWTGN is upper-bounded by the weighted nuclear norm, i.e., $\|\mathcal{Y}\|_{\Gamma,\Lambda}\leqslant\|\mathcal{Y}\|_{\Lambda,\ast}.$
	\\\textbf{(d) Asymptotic nuclear norm property}: The BEWTGN approaches the weighted nuclear norm asymptotically, i.e., $\lim\limits_{\Gamma\to\infty}\|\mathcal{Y}\|_{\Gamma,\Lambda}=\|\mathcal{Y}\|_{\Lambda,\ast}.$
	\\\textbf{(e) Unitary invariance}: The BEWTGN is unitary invariant, i.e., $\|\mathcal{U}\ast\mathcal{Y}\ast\mathcal{V}\|_{\Gamma,\Lambda}=\|\mathcal{Y}\|_{\Gamma,\Lambda}$, for unitary tensor $\mathcal{U}$ and $\mathcal{V}$.
\label{proposition1}\end{proposition}
\begin{IEEEproof}
		Let
	\begin{eqnarray}
	p(\mathcal{Y})=\|\mathcal{Y}\|_{\frac{\upsilon}{\Gamma},\ast}+\frac{1}{2}\|\frac{\upsilon-\Lambda\star\Gamma}{\Gamma\star\Gamma}\|_{F}^{2}.\nonumber
	\end{eqnarray} 
	\\(a) Since $p(\mathcal{Y})$ is the sum of two non-negative functions, $\|\mathcal{Y}\|_{\Gamma,\Lambda}\geqslant0$. The equality holds if $\|\mathcal{Y}\|_{\frac{\upsilon}{\Gamma},\ast}=0$, i.e., $\mathcal{Y}=\textbf{0}$ or $\upsilon=\textbf{0}$, the latter being the trivial solution.
	\\(b) The function $p(\mathcal{Y})$ is separable of $\mathcal{Y}$, i.e.,
	\begin{eqnarray}
	p(\mathcal{Y})&=&\sum_{i_{3}=1}^{\mathit{I}_{3}}\sum_{j=1}^{R}\frac{\upsilon_{(i_{3},j)}}{\Gamma_{(i_{3},j)}}\sigma_{j}(\bar{\mathcal{Y}}^{(i_{3})})+\frac{(\upsilon_{(i_{3},j)}-\Lambda_{(i_{3},j)}\Gamma_{(i_{3},j)})^{2}}{2\Gamma_{(i_{3},j)}},\nonumber
	\\&=&\sum_{i_{3}=1}^{\mathit{I}_{3}}\sum_{j=1}^{R}p(\sigma_{j}(\bar{\mathcal{Y}}^{(i_{3})}))\nonumber
	\end{eqnarray}
	Since $p$ is an affine function in $\sigma_{j}(\bar{\mathcal{Y}}^{(i_{3})})$, we can write, for $0\leqslant\alpha\leqslant1$, that
	\begin{eqnarray}
	&&\|(\alpha\mathcal{Y}_{1}+(1-\alpha)\mathcal{Y}_{2})\|_{\Gamma,\Lambda}\nonumber
	\\&&=\min_{\upsilon}p(\alpha\mathcal{Y}_{1}+(1-\alpha)\mathcal{Y}_{2})\nonumber
	\\&&\geqslant\min_{\upsilon}\alpha p(\mathcal{Y}_{1})+(1-\alpha)p(\mathcal{Y}_{2})\nonumber
	\\&&\geqslant\min_{\upsilon}\alpha p(\mathcal{Y}_{1})+\min_{\upsilon}(1-\alpha)p(\mathcal{Y}_{2})\nonumber
	\\&&=\alpha \|\mathcal{Y}_{1}\|_{\Gamma,\Lambda}+(1-\alpha)\|\mathcal{Y}_{2}\|_{\Gamma,\Lambda}.\nonumber
	\end{eqnarray}
	Hence, $\|\mathcal{Y}\|_{\Gamma,\Lambda}$ is concave in the modulus of the singular values of $\mathcal{Y}$.
	\\(c) For $\gamma>1$,
	\begin{eqnarray}
	\|\mathcal{Y}\|_{\Gamma,\Lambda}\leqslant\|\mathcal{Y}\|_{\frac{\upsilon}{\Gamma},\ast}+\frac{1}{2}\|\frac{\upsilon-\Lambda\star\Gamma}{\Gamma\star\Gamma}\|_{F}^{2},\quad \forall\,\upsilon\in\mathbb{R}^{\mathit{I}_{3}\times R}_{\geqslant0}.\nonumber
	\end{eqnarray}
	Therefore, the above inequality holds true even for $\upsilon=\Lambda\star\Gamma$, which leads to $\|\mathcal{Y}\|_{\Gamma,\Lambda}\leqslant\|\mathcal{Y}\|_{\frac{\upsilon}{\Gamma},\ast}$.
	\\(d) As $\Gamma\to\infty$, the optimal $\upsilon$ in (\ref{ewtgn}) is $\Lambda\star\Gamma$. Therefore, $\lim\limits_{\Gamma\to\infty}\|\mathcal{Y}\|_{\Gamma,\Lambda}=\|\mathcal{Y}\|_{\frac{\upsilon}{\Gamma},\ast}$.
	\\(e) Consider the definition of the tensor weighted nuclear norm
	\begin{eqnarray}
	\|\mathcal{Y}\|_{\frac{\upsilon}{\Gamma},\ast}:=\sum_{i=1}^{\mathit{I}_{3}}\sum_{j=1}^{R}\frac{\upsilon_{(i,j)}}{\Gamma_{(i,j)}}\sigma_{j}(\bar{\mathcal{Y}}^{(i_{3})}),
	\end{eqnarray}
	where $\sigma(\bar{\mathcal{Y}}^{(i)})$ denote the singular values of $i$th slice of $\bar{\mathcal{Y}}$. Let $w=\frac{\upsilon}{\Gamma}$, we can experss
	\begin{eqnarray}
	\|\mathcal{Y}\|_{w,\ast}=\sum_{i=1}^{\mathit{I}_{3}}Tr(diag(w(i,:))\sqrt{\bar{\mathcal{Y}}^{(i)T}\bar{\mathcal{Y}}^{(i)}}),
	\end{eqnarray}
	where $diag(w(i,:))$ denotes a diagonal matrix with diagonal entries coming from the elements of $i$th column vector of $w$, and $Tr(\cdot)$ denotes the Trace operator. 
	
	Next, consider $\|\mathcal{U}\ast \mathcal{Y}\ast\mathcal{V}\|_{w,\ast}$, where $\mathcal{U}$ and $\mathcal{V}$ are unitary tensors:
	\begin{eqnarray}
	\|\mathcal{U}\ast\mathcal{Y}\ast\mathcal{V}\|_{w,\ast}:=\sum_{i=1}^{\mathit{I}_{3}}\|\bar{\mathcal{U}}^{(i)}\bar{\mathcal{Y}}^{(i)}\bar{\mathcal{V}}^{(i)}\|_{w(i,:),\ast}\nonumber
	\\=\sum_{i=1}^{\mathit{I}_{3}}Tr(diag(w(i,:))\sqrt{(\bar{\mathcal{U}}^{(i)}\bar{\mathcal{Y}}^{(i)}\bar{\mathcal{V}}^{(i)})^{T}(\bar{\mathcal{U}}^{(i)}\bar{\mathcal{Y}}^{(i)}\bar{\mathcal{V}}^{(i)})}.
	\end{eqnarray}
	Properties of the tensor generated by performing discrete Fourier transformation (DFT) show that $\bar{\mathcal{U}}^{(i)}$ and $\bar{\mathcal{V}}^{(i)}$ are unitary matrices. Then, we get the following formula:
	\begin{eqnarray}
	\|\mathcal{U}\ast\mathcal{Y}\ast\mathcal{V}\|_{w,\ast}=\|\mathcal{Y}\|_{w,\ast}.\label{Unitary}
	\end{eqnarray}
	Based on (\ref{Unitary}), we can express
	\begin{eqnarray}
	&&\|\mathcal{U}\ast\mathcal{Y}\ast\mathcal{V}\|_{\Gamma,\Lambda}\nonumber
	\\&&=\min_{\upsilon} \left\lbrace \|\mathcal{U}\ast\mathcal{Y}\ast\mathcal{V}\|_{\frac{\upsilon}{\Gamma},\ast}+\frac{1}{2}\|\frac{\upsilon-\Lambda\star\Gamma}{\Gamma\star\Gamma}\|_{F}^{2}\right\rbrace \nonumber
	\\&&=\min_{\upsilon} \left\lbrace \|\mathcal{Y}\|_{\frac{\upsilon}{\Gamma},\ast}+\frac{1}{2}\|\frac{\upsilon-\Lambda\star\Gamma}{\Gamma\star\Gamma}\|_{F}^{2}\right\rbrace \nonumber
	\\&&=\|\mathcal{Y}\|_{\Gamma,\Lambda}\nonumber,
	\end{eqnarray}
	which is the BEWTGN. This establishes the unitary  invariance of the BEWTGN.
\end{IEEEproof}

\section{PROXIMAL OPERATORS FOR THE BEMCP THEOREM and BEWTGN}
In this section, to solve the model established based on the BEMCP theorem, we present the proximal operators for the BEMCP theorem and BEWTGN.
\begin{theorem}[Proximal operator for the BEMCP]
	Consider the BEMCP given in (\ref{BEMCP}). Its proximal operator denoted by $\mathit{P}_{\gamma,\lambda}:\mathbb{R}\to\mathbb{R},\gamma\in\mathbb{R}_{>1}$, $\lambda\in\mathbb{R}_{\geqslant0}$ and defined as follows:
	\begin{eqnarray}
	\mathit{P}_{\gamma,\lambda}(y)=\arg\min_{g}\left\lbrace\frac{1}{2}(g-y)^{2}+h_{\gamma,\lambda}(g)\right\rbrace ,\label{temcp1}
	\end{eqnarray}
	is given by
	\begin{eqnarray}
	\mathit{P}_{\gamma,\lambda}(y)=\min \left( \rvert y\rvert,\max\left( \frac{\gamma(\rvert y\rvert-{\lambda})}{\gamma-1},0\right) \right)  sign(y).\label{temcp2}
	\end{eqnarray}
\label{thTEMCP}\end{theorem}
\begin{IEEEproof}
	Let 
	\begin{eqnarray}
	&&\mathit{P}_{\gamma,\lambda}(y)\nonumber
	\\&&=\arg\min_{g} \left\lbrace \frac{1}{2}(g-y)^{2}+\min_{\upsilon\geqslant0}\left\lbrace \frac{2\upsilon\rvert g \rvert+(\upsilon-\lambda\gamma)^{2}}{2\gamma}\right\rbrace \right\rbrace ,\nonumber
	\\&&=\arg\min_{g} \left\lbrace \frac{1}{2}(g-y)^{2}+h(g,\upsilon^{\ast})\right\rbrace ,\nonumber
	\end{eqnarray}
	where
	\begin{eqnarray}
	\upsilon^{\ast}=\arg\min_{\upsilon\geqslant0}h(g,\upsilon)=max(\lambda\gamma-\vert g\vert,0).\nonumber
	\end{eqnarray}
	We must now determine $\mathit{P}_{\gamma,\lambda}(y)$ for a given $\upsilon^{\ast}$. This is derived considering various values that $\upsilon^{\ast}$ can take:
	\\(1) Case1: $\upsilon^{\ast}=0$. Correspondingly, $h(g,\upsilon^{\ast})=\frac{\lambda^{2}\gamma}{2}$, and then
	\begin{eqnarray}
	\mathit{P}_{\gamma,\lambda}(y)=\arg\min_{g}\left\lbrace\frac{1}{2}(g-y)^{2}+\frac{\lambda^{2}\gamma}{2} \right\rbrace =y.\nonumber
	\end{eqnarray}
	The condition $\upsilon^{\ast}=0$ translates to $\vert g \vert\geqslant\lambda\gamma$, therefore, 
	\begin{eqnarray}
	\mathit{P}_{\gamma,\lambda}(y)=y,\quad\text{for}\quad \vert g \vert\geqslant\lambda\gamma.\label{case1}
	\end{eqnarray}
	\\(2) Case2: $0<\upsilon^{\ast}<\lambda\gamma$. By definition of $\upsilon^{\ast}$, we have that
	\begin{eqnarray}
	\upsilon^{\ast}=\lambda\gamma-\vert g\vert,\quad \vert g\vert\leqslant\lambda\gamma.
	\end{eqnarray}
	Further, considering $0<g\leqslant\lambda\gamma$ gives $\upsilon^{\ast}=\lambda\gamma-g$, which result in 
	\begin{eqnarray}
	\mathit{P}_{\gamma,\lambda}(y)&=&\arg\min_{g} \left\lbrace \frac{1}{2}(g-y)^{2}+\frac{2\upsilon^{\ast}g +(\upsilon^{\ast}-\lambda\gamma)^{2}}{2\gamma}\right\rbrace ,\nonumber
	\\&=& \frac{\gamma}{\gamma-1}(y-\lambda)\nonumber\quad \text{for}\quad\lambda<y<\lambda\gamma. 
	\end{eqnarray}
	Similarly, considering $-\lambda\gamma\leqslant g<0$ gives $\upsilon^{\ast}=\lambda\gamma+g$, which result in 
	\begin{eqnarray}
	\mathit{P}_{\gamma,\lambda}(y)&=&\arg\min_{g} \left\lbrace \frac{1}{2}(g-y)^{2}+\frac{2\upsilon^{\ast}(-g) +(\upsilon^{\ast}-\lambda\gamma)^{2}}{2\gamma}\right\rbrace ,\nonumber
	\\&=& \frac{\gamma}{\gamma-1}(-y-\lambda)\nonumber\quad \text{for}\quad-\lambda\gamma<y<-\lambda. 
	\end{eqnarray}
	Combining the two expressions brings
	\begin{eqnarray}
	\mathit{P}_{\gamma,\lambda}(y)=\frac{\gamma}{\gamma-1}(\vert y\vert-\lambda)sign(y),\quad \text{for}\quad\lambda<\vert y\vert<\lambda\gamma. \label{case2}
	\end{eqnarray}
	\\(3) Case3: $\upsilon^{\ast}=\lambda\gamma$. This induces $g=0$, which implies $\mathit{P}_{\gamma,\lambda}(y)=0$. Substituting $\upsilon^{\ast}=\lambda\gamma$ in the definition of $\mathit{P}_{\gamma,\lambda}(y)$ gives 
	\begin{eqnarray}
	0=\arg\min_{g} \left\lbrace \frac{1}{2}(g-y)^{2}+\lambda\vert g\vert\right\rbrace .\nonumber
	\end{eqnarray}
	Considering the subdifferential, we get $0\in -y+\lambda\partial\vert g\vert$, or equivalently, $\vert y\vert\leqslant\lambda$. Therefore,
	\begin{eqnarray}
	\mathit{P}_{\gamma,\lambda}(y)=0, \quad\text{for}\quad \vert y\vert\leqslant\lambda\label{case3}.
	\end{eqnarray}
	Combining equations (\ref{case1}), (\ref{case2}), and (\ref{case3}) yields
	\begin{eqnarray}
	\mathit{P}_{\gamma,\lambda}(y)=\min \left( \rvert y\rvert,\max \left( \frac{\gamma(\rvert y\rvert-{\lambda})}{\gamma-1},0\right)  \right) sign(y).
	\end{eqnarray}
%	Consolidating the solutions of the scalar sub-problems results in (\ref{temcp2}).
\end{IEEEproof}
\begin{theorem}[Proximal operator for the BEWTGN]
	Consider the BEWTGN given in (\ref{ewtgn}). Its proximal operator denoted by $\mathit{S}_{\Gamma,\Lambda}:\mathbb{R}^{\mathit{I}_{1}\times\mathit{I}_{2}\times\mathit{I}_{3}}\to\mathbb{R}^{\mathit{I}_{1}\times\mathit{I}_{2}\times\mathit{I}_{3}},\Gamma\in\mathbb{R}^{R\times\mathit{I}_{3}}_{>1}$, $\bar{\Lambda}\in\mathbb{R}^{R\times\mathit{I}_{3}}_{\geqslant0}$, $R=\min\{\mathit{I}_{1},\mathit{I}_{2}\}$ and defined as follows:
	\begin{eqnarray}
	\mathit{S}_{\Gamma,\Lambda	}(\mathcal{Y})=\arg\min_{\mathcal{L}} \left\lbrace \frac{1}{2}\|\mathcal{L}-\mathcal{Y}\|_{F}^{2}+\|\mathcal{L}\|_{\Gamma,\Lambda}\right\rbrace ,
	\end{eqnarray}
	is given by
	\begin{eqnarray}
	\mathit{S}_{\Gamma,\Lambda}(\mathcal{Y})=\mathcal{U}\ast\mathcal{S}_{1}\ast\mathcal{V}^{H}.\label{opewtgn}
	\end{eqnarray}
	where $\mathcal{U}$ and $\mathcal{V}$ are derived from the t-SVD of $\mathcal{Y}=\mathcal{U}\ast\mathcal{S}_{2}\ast\mathcal{V}^{H}$. More importantly, the $i$th front slice of DFT of $\mathcal{S}_{1}$ and $\mathcal{S}_{2}$, i.e., $\bar{\mathcal{S}}^{(i)}_{1}=\sigma(\bar{\mathcal{L}}^{(i)})$ and $\bar{\mathcal{S}}^{(i)}_{2}=\sigma(\bar{\mathcal{Y}}^{(i)})$, has the following relationship:
	\begin{eqnarray}
	\sigma(\bar{\mathcal{L}}^{(i)})=\min \left\lbrace \sigma(\bar{\mathcal{Y}}^{(i)}),\max \left\lbrace \dfrac{\Gamma_{(i,:)}(\sigma(\bar{\mathcal{Y}}^{(i)})-\Lambda_{(i,:)})}{\Gamma_{(i,:)}-1},0\right\rbrace \right\rbrace .\nonumber
	\end{eqnarray}
%	where $\tau=\rho$.
\label{thEWTGN}\end{theorem}
\begin{IEEEproof}
	Let $\mathcal{Y}=\mathcal{U}\ast\mathcal{S}_{2}\ast\mathcal{V}^{H}$ and $\mathcal{L}=\mathcal{W}\ast\mathcal{S}_{1}\ast\mathcal{R}^{H}$ be the t-SVD of $\mathcal{Y}$ and $\mathcal{L}$, respectively. Consider
	\begin{eqnarray}
	\mathit{S}_{\Gamma,\Lambda}(\mathcal{Y})&=&\arg\min_{\mathcal{L}}\frac{1}{2}\|\mathcal{L}-\mathcal{Y}\|_{F}^{2}+\|\mathcal{L}\|_{\Gamma,\Lambda}\nonumber
	\\&=&\arg\min_{\mathcal{L}}\frac{1}{2}\|\mathcal{W}\ast\mathcal{S}_{1}\ast\mathcal{R}^{H}-\mathcal{U}\ast\mathcal{S}_{2}\ast\mathcal{V}^{H}\|_{F}^{2}\nonumber
	\\&&+\|\mathcal{L}\|_{\Gamma,\Lambda}\nonumber
	\\&=&\arg\min_{\bar{\mathcal{L}}}\sum_{i=1}^{\mathit{I}_{3}}\frac{1}{2}\|\bar{\mathcal{W}}^{(i)}\ast\bar{\mathcal{S}_{1}}^{(i)}\ast\bar{\mathcal{R}}^{(i)H}-\bar{\mathcal{U}}^{(i)}\nonumber
	\\&&\ast\bar{\mathcal{S}_{2}}^{(i)}\ast\bar{\mathcal{V}}^{(i)H}\|_{F}^{2}+\|\bar{\mathcal{L}}^{(i)}\|_{\Gamma_{(i,:)},\Lambda_{(i,:)}}.\label{profewtg}
	\end{eqnarray}
	It can be found that (\ref{profewtg}) is separable and can be divided into $\mathit{I}_{3}$ sub-problems. For the $i$th sub-problem:
	\begin{eqnarray}
	&&\arg\min_{\bar{\mathcal{L}}^{(i)}}\frac{1}{2}\|\bar{\mathcal{W}}^{(i)}\ast\bar{\mathcal{S}_{1}}^{(i)}\ast\bar{\mathcal{R}}^{(i)H}-\bar{\mathcal{U}}^{(i)}\ast\bar{\mathcal{S}_{2}}^{(i)}\ast\bar{\mathcal{V}}^{(i)H}\|_{F}^{2}\nonumber
	\\&&+\|\bar{\mathcal{L}}^{(i)}\|_{\Gamma_{(i,:)},\Lambda_{(i,:)}}\nonumber
	\\&&=\arg\min_{\bar{\mathcal{L}}^{(i)}}\frac{1}{2}Tr(\bar{\mathcal{S}_{1}}^{(i)}\bar{\mathcal{S}_{1}}^{(i)H})+\frac{1}{2}Tr(\bar{\mathcal{S}_{2}}^{(i)}\bar{\mathcal{S}_{2}}^{(i)H})\nonumber
	\\&&+ Tr(\bar{\mathcal{L}}^{(i)H}\bar{\mathcal{Y}}^{(i)})+\|\bar{\mathcal{L}}^{(i)}\|_{\Gamma_{(i,:)},\Lambda_{(i,:)}}.
	\end{eqnarray}
	Invoking von Neumann's trace inequality \cite{mirsky1975trace}, we can write
	\begin{eqnarray}
	&&\arg\min_{\bar{\mathcal{L}}^{(i)}}\frac{1}{2}\|\bar{\mathcal{W}}^{(i)}\ast\bar{\mathcal{S}_{1}}^{(i)}\ast\bar{\mathcal{R}}^{(i)H}-\bar{\mathcal{U}}^{(i)}\ast\bar{\mathcal{S}_{2}}^{(i)}\ast\bar{\mathcal{V}}^{(i)H}\|_{F}^{2}\nonumber
	\\&&+\|\bar{\mathcal{L}}^{(i)}\|_{\Gamma_{(i,:)},\Lambda_{(i,:)}}\nonumber
	\\&&\geqslant\arg\min_{\bar{\mathcal{S}_{1}}^{(i)}}\frac{1}{2}Tr(\bar{\mathcal{S}_{1}}^{(i)}\bar{\mathcal{S}_{1}}^{(i)H})+\frac{1}{2}Tr(\bar{\mathcal{S}_{2}}^{(i)}\bar{\mathcal{S}_{2}}^{(i)H})\nonumber
	\\&&+Tr(\bar{\mathcal{S}_{2}}^{(i)}\bar{\mathcal{S}_{1}}^{(i)H})+\|\bar{\mathcal{L}}^{(i)}\|_{\Gamma_{(i,:)},\Lambda_{(i,:)}}\nonumber\nonumber
	\\&&=\arg\min_{\sigma(\bar{\mathcal{L}}^{(i)})}\frac{1}{2}\|\sigma(\bar{\mathcal{L}}^{(i)})-\sigma(\bar{\mathcal{Y}}^{(i)})\|_{F}^{2}+\|\bar{\mathcal{L}}^{(i)}\|_{\Gamma_{(i,:)},\Lambda_{(i,:)}}.\nonumber
	\end{eqnarray}
	The equality holds when $\bar{\mathcal{W}}^{(i)}=\bar{\mathcal{U}}^{(i)}$ and $\bar{\mathcal{R}}^{(i)}=\bar{\mathcal{V}}^{(i)}$. Therefore, the optimal solution to (\ref{profewtg}) is obtained by solving the
	problem given below:
	\begin{eqnarray}
	\sigma(\bar{\mathcal{L}}^{(i)})=\min \left\lbrace \sigma(\bar{\mathcal{Y}}^{(i)}),\max \left\lbrace \dfrac{\Gamma_{(i,:)}(\sigma(\bar{\mathcal{Y}}^{(i)})-\Lambda_{(i,:)})}{\Gamma_{(i,:)}-1},0\right\rbrace \right\rbrace .\nonumber
	\end{eqnarray}
\end{IEEEproof}

%The proximal operators are crucial in deriving the LRTC algorithms that will be developed in the following section.

\section{THE BEMCP MODELS AND SOLVING ALGORITHMS}
In this section, in order to better verify the superiority of our BEMCP theorem, before giving the BEMCP model based on the N-tubal rank, we first review the models of traditional MCP and EMCP based on the N-tubal rank. And the traditional MCP model based on N-tubal rank is called the NCMP model.
\begin{eqnarray}
\min_{\mathcal{X}}\sum_{1\leqslant l_{1}<l_{2}\leqslant N}^{N}\alpha_{l_{1}l_{2}}\|\mathcal{X}_{(l_{1}l_{2})}\|_{\ast}\quad s.t.\quad \mathcal{P}_{\Omega}(\mathcal{X}-\mathcal{Z})=\mathbf{0},\label{LRTC}
\end{eqnarray}
Applying different MCP non-convex functions to singular values of problem (\ref{LRTC}) leads to the following model.
\subsection{The NMCP model}
Using the MCP function, we get the following optimization model:
\begin{eqnarray}
\min_{\mathcal{X}}\sum_{1\leqslant l_{1}<l_{2}\leqslant N}^{N}\alpha_{l_{1}l_{2}}\|\mathcal{X}_{(l_{1}l_{2})}\|_{\gamma,\lambda}\quad s.t.\quad \mathcal{P}_{\Omega}(\mathcal{X}-\mathcal{Z})=\mathbf{0}.\label{NMCP}
\end{eqnarray}
Under the framework of alternation direction method of multipliers (ADMM) \cite{683201156}, \cite{6122011620},  \cite{lin2010augmented}, the easy-to-implement optimization strategy could be provided to solve (\ref{NMCP}). We introduce a set of tensors $\{\mathcal{Y}_{l_{1}l_{2}}=\mathcal{X}\}_{1\leqslant l_{1}<l_{2}\leqslant N}^{N}$ and transfer optimization problem (\ref{NMCP}), in its augmented Lagrangian form, as follows:
\begin{eqnarray}
&&J(\mathcal{X},\mathcal{Y},\mathcal{Q})\nonumber
\\&&=\sum_{1\leqslant l_{1}<l_{2}\leqslant N}^{N}\alpha_{l_{1}l_{2}}\|\mathcal{Y}_{l_{1}l_{2}(l_{1}l_{2})}\|_{\gamma,\lambda}\nonumber
+\frac{\rho_{l_{1}l_{2}}}{2}\|\mathcal{X}-\mathcal{Y}_{l_{1}l_{2}}+\frac{\mathcal{Q}_{l_{1}l_{2}}}{\rho_{l_{1}l_{2}}}\|_{F}^{2} \nonumber
\\&&\quad s.t.\quad \mathcal{P}_{\Omega}(\mathcal{X}-\mathcal{Z})=\mathbf{0},\label{LRTC1}
\end{eqnarray}
where $\{\mathcal{Q}_{l_{1}l_{2}}\}_{1\leqslant l_{1}<l_{2}\leqslant N}^{N}$ are tensor Lagrangian multiplier sets; $\{\rho_{l_{1}l_{2}}\}_{1\leqslant l_{1}<l_{2}\leqslant N}^{N}>0$ are the augmented Lagrangian parameters; $\alpha_{l_{1}l_{2}}\geqslant0$ are N-tubal rank weights and $\sum_{1\leqslant l_{1}<l_{2}\leqslant N}^{N}\alpha_{l_{1}l_{2}}=1$. Besides, variables $\mathcal{X}$, $\mathcal{Y}$, $\mathcal{Q}$ are updated alternately in the order of $\mathcal{Y}\to\mathcal{X}\to\mathcal{Q}$. For convenience, we mark the updated variable as $(\cdot)^{+}$. The update equations are acquired in the following. 

\textbf{Update $\mathcal{Y}$}: Fix other variables, and the corresponding optimization is as follows: 
\begin{eqnarray}
	\mathcal{Y}_{l_{1}l_{2}}^{+}=\arg\min_{\mathcal{Y}_{l_{1}l_{2}}}\alpha_{l_{1}l_{2}}\|\mathcal{Y}_{l_{1}l_{2}(l_{1}l_{2})}\|_{\gamma,\lambda}+\frac{\rho_{l_{1}l_{2}}}{2}\|\mathcal{X}-\mathcal{Y}_{l_{1}l_{2}}+\frac{\mathcal{Q}_{l_{1}l_{2}}}{\rho_{l_{1}l_{2}}}\|_{F}^{2}. \nonumber
\end{eqnarray}
Calling Theorem \ref{thEWTGN}, the solution to the above optimization is given by:
\begin{eqnarray}
	\mathcal{Y}_{l_{1}l_{2}}^{+}=\mathit{S}_{\frac{\gamma\rho_{l_{1}l_{2}}}{\alpha_{l_{1}l_{2}}},\frac{\lambda\rho_{l_{1}l_{2}}}{\alpha_{l_{1}l_{2}}}}(\mathcal{X}+\frac{\mathcal{Q}_{l_{1}l_{2}}}{\rho_{l_{1}l_{2}}}),\label{UP1Y}
\end{eqnarray}
where $\mathit{S}$ denotes the proximal operator defined in (\ref{opewtgn}).

\textbf{Update $\mathcal{X}$}: The closed form of $\mathcal{X}$ can be acquired by setting the derivative of (\ref{LRTC1}) to zero. We can now update $\mathcal{X}$ by the following equation:
\begin{eqnarray}
\mathcal{X}^{+}=\mathcal{P}_{\Omega^{c}}(\frac{\sum_{1\leqslant l_{1}< l_{2}\leqslant N}\rho_{l_{1}l_{2}}(\mathcal{Y}_{l_{1}l_{2}}-\frac{\mathcal{Q}_{l_{1}l_{2}}}{\rho_{l_{1}l_{2}}})}{\sum_{1\leqslant l_{1}< l_{2}\leqslant N}\rho_{l_{1}l_{2}}})+\mathcal{P}_{\Omega}(\mathcal{Z}).\label{UP1X}
\end{eqnarray}

\textbf{Update $\mathcal{Q}$}: Finally, multipliers $\mathcal{Q}_{l_{1}l_{2}}$ are updated as follows:
\begin{eqnarray}
\mathcal{Q}_{l_{1}l_{2}}^{+}=\mathcal{Q}_{l_{1}l_{2}}+\rho_{l_{1}l_{2}}(\mathcal{X}-\mathcal{Y}_{l_{1}l_{2}}).\label{UP1Q}
\end{eqnarray}
The optimization steps of the NMCP formulation are listed in Algorithm \ref{TC1}.
\begin{algorithm}[!h]
	\caption{NMCP} %算法的名字
	\hspace*{0.02in} {\bf Input:} %算法的输入,  \hspace*{0.02in}用来控制位置, 同时利用 \\ 进行换行
	An incomplete tensor $\mathcal{Z}$, the index set of the known elements $\Omega$, convergence criteria $\epsilon$, maximum iteration number $K$. \\
	\hspace*{0.02in} {\bf Initialization:} %算法的结果输出
	$\mathcal{X}^{0}=\mathcal{Z}_{\Omega}$, $\mathcal{Y}_{l_{1}l_{2}}^{0}=\mathcal{X}^{0}$, $\rho_{l_{1}l_{2}}^{0}>0$, $\mu>1$.
	\begin{algorithmic}
		%\State some description % \State 后写一般语句		
		\While{not converged and $k<K$} % While语句, 需要和EndWhile对应
		%\For{k=1,2,$\cdots$,K} % For 语句, 需要和EndFor对应	
		\State Updating $\mathcal{Y}_{l_{1}l_{2}}^{k}$ via (\ref{UP1Y});
		\State Updating $\mathcal{X}^{k}$ via (\ref{UP1X});
		\State Updating the multipliers $\mathcal{Q}_{l_{1}l_{2}}^{k}$ via (\ref{UP1Q});
		\State $\rho_{l_{1}l_{2}}^{k}=\mu\rho_{l_{1}l_{2}}^{k-1}$, $k=k+1$;
		\State Check the convergence conditions $\|\mathcal{X}^{k+1}-\mathcal{X}^{k}\|_{\infty}\leq\epsilon$.
		%	\EndFor
		\EndWhile
		\State \Return $\mathcal{X}^{k+1}$.
	\end{algorithmic}
	\hspace*{0.02in} {\bf Output:} %算法的结果输出
	Completed tensor $\mathcal{X}=\mathcal{X}^{k+1}$.
	\label{TC1}\end{algorithm}

\subsection{The EMCP model}
Similarly, using the EMCP Theorem, we get the following optimization model:
\begin{eqnarray}
\min_{\mathcal{X}}\sum_{1\leqslant l_{1}<l_{2}\leqslant N}^{N}\alpha_{l_{1}l_{2}}\|\mathcal{X}_{(l_{1}l_{2})}\|_{\gamma,\bar{\Lambda}_{(l_{1}l_{2})}}\quad s.t. \mathcal{P}_{\Omega}(\mathcal{X}-\mathcal{Z})=\mathbf{0}.\label{EMCP}
\end{eqnarray}
Under the framework of the alternation direction method of multipliers (ADMM), the easy-to-implement optimization strategy could be provided to solve (\ref{EMCP}). We introduce a set of tensors $\{\mathcal{Y}_{l_{1}l_{2}}=\mathcal{X}\}_{1\leqslant l_{1}<l_{2}\leqslant N}^{N}$ and transfer optimization problem (\ref{EMCP}), in its augmented Lagrangian form, as follows:
\begin{eqnarray}
&&J(\mathcal{X},\mathcal{Y}, \bar{\Lambda}, W,\mathcal{Q})\nonumber
\\&&=\sum_{1\leqslant l_{1}<l_{2}\leqslant N}^{N}\alpha_{l_{1}l_{2}}\|\mathcal{Y}_{l_{1}l_{2}(l_{1}l_{2})}\|_{\gamma,\bar{\Lambda}_{l_{1}l_{2}}}\label{EMCP2}\\&&+\frac{\rho_{l_{1}l_{2}}}{2}\|\mathcal{X}-\mathcal{Y}_{l_{1}l_{2}}+\frac{\mathcal{Q}_{l_{1}l_{2}}}{\rho_{l_{1}l_{2}}}\|_{F}^{2} 
\quad s.t.\quad \mathcal{P}_{\Omega}(\mathcal{X}-\mathcal{Z})=\mathbf{0}\nonumber
\end{eqnarray}
where $\mathcal{Y}$ and $\mathcal{Q}$ are tensor sets; $\bar{\Lambda}$ and $W$ are matrix sets; $\{\mathcal{Y}_{l_{1}l_{2}}=\mathcal{X}\}_{1\leqslant l_{1}<l_{2}\leqslant N}^{N}$; $\{\mathcal{Q}_{l_{1}l_{2}}\}_{1\leqslant l_{1}<l_{2}\leqslant N}^{N}$ are Lagrangian multipliers; $\{\bar{\Lambda},W\}_{1\leqslant l_{1}<l_{2}\leqslant N}^{N}\in\mathbb{R}^{\mathit{I}_{3}\times R}$ are MCP variable and weight sets, respectively; $\{\rho_{l_{1}l_{2}}\}_{1\leqslant l_{1}<l_{2}\leqslant N}^{N}>0$ are the augmented Lagrangian parameters; $\alpha_{l_{1}l_{2}}\geqslant0$ are weights and $\sum_{1\leqslant l_{1}<l_{2}\leqslant N}^{N}\alpha_{l_{1}l_{2}}=1$.

Besides, variables $\mathcal{X}$, $\mathcal{Y}$, $\bar{\Lambda}$, $W$, $\mathcal{Q}$ are updated alternately in the order of $\mathcal{Y}\to W\to \bar{\Lambda}\to\mathcal{X}\to\mathcal{Q}$. The update equations are acquired in the following. 

\textbf{Update $\mathcal{Y}$}: Fix other variables, and the corresponding optimization is as follows: 
\begin{eqnarray}
\mathcal{Y}_{l_{1}l_{2}}^{+}&=&\arg\min_{\mathcal{Y}_{l_{1}l_{2}}}\alpha_{l_{1}l_{2}}\|\mathcal{Y}_{l_{1}l_{2}(l_{1}l_{2})}\|_{\gamma,\bar{\Lambda}_{l_{1}l_{2}}}\nonumber
\\&&+\frac{\rho_{l_{1}l_{2}}}{2}\|\mathcal{X}-\mathcal{Y}_{l_{1}l_{2}}+\frac{\mathcal{Q}_{l_{1}l_{2}}}{\rho_{l_{1}l_{2}}}\|_{F}^{2}. \nonumber
\end{eqnarray}
Invoking Theorem \ref{thEWTGN}, the solution to the above optimization is given by:
\begin{eqnarray}
\mathcal{Y}_{l_{1}l_{2}}^{+}=\mathit{S}_{\frac{\gamma\rho_{l_{1}l_{2}}}{\alpha_{l_{1}l_{2}}},\frac{\bar{\Lambda}_{l_{1}l_{2}}\rho_{l_{1}l_{2}}}{\alpha_{l_{1}l_{2}}}}(\mathcal{X}+\frac{\mathcal{Q}_{l_{1}l_{2}}}{\rho_{l_{1}l_{2}}}),\label{UP2Y}
\end{eqnarray}
where $\mathit{S}$ denotes the proximal operator defined in (\ref{opewtgn}).

\textbf{Update $W$}:  Retaining only those components in $\mathcal{Y}_{l_{1}l_{2}}$ in ($\ref{EMCP2}$) that depend on $W_{l_{1}l_{2}}$, we write
\begin{eqnarray}
W_{l_{1}l_{2}}^{+}=\arg\min_{W_{l_{1}l_{2}}}\sum_{i_{3}=1}^{\mathit{I}_{3}}\|\bar{\mathcal{Y}}_{l_{1}l_{2}}^{(i_{3})}\|_{W_{l_{1}l_{2}(i_{3},:)},\ast}+\frac{\gamma}{2}\|W_{l_{1}l_{2}}-\bar{\Lambda}_{l_{1}l_{2}}\|_{F}^{2},\nonumber
\end{eqnarray}
which has the following closed-form solution:
\begin{eqnarray}
W_{l_{1}l_{2}(i_{3},:)}^{+}=\max(\bar{\Lambda}_{l_{1}l_{2}(i_{3},:)}-\frac{\sigma(\mathcal{Y}_{l_{1}l_{2}}^{+(i_{3})})}{\gamma},0).\label{UP2W}
\end{eqnarray}

\textbf{Update $\bar{\Lambda}$}: The update for $\bar{\Lambda}_{l_{1}l_{2}}$ has the following closed-form solution:
\begin{eqnarray}
\bar{\Lambda}_{l_{1}l_{2}}^{+}=\arg\min_{\bar{\Lambda}_{l_{1}l_{2}}}\|W_{l_{1}l_{2}}-\bar{\Lambda}_{l_{1}l_{2}}\|_{F}^{2}=W_{l_{1}l_{2}}^{+}.\label{UP2LA}
\end{eqnarray}

\textbf{Update $\mathcal{X}$}: The closed form of $\mathcal{X}$ can be acquired by setting the derivative of (\ref{EMCP2}) to zero. We can now update $\mathcal{X}$ by the following equation:
\begin{eqnarray}
\mathcal{X}^{+}=\mathcal{P}_{\Omega^{c}}(\frac{\sum_{1\leqslant l_{1}< l_{2}\leqslant N}\rho_{l_{1}l_{2}}(\mathcal{Y}_{l_{1}l_{2}}-\frac{\mathcal{Q}_{l_{1}l_{2}}}{\rho_{l_{1}l_{2}}})}{\sum_{1\leqslant l_{1}< l_{2}\leqslant N}\rho_{l_{1}l_{2}}})+\mathcal{P}_{\Omega}(\mathcal{Z}).\label{UP2X}
\end{eqnarray}

\textbf{Update $\mathcal{Q}$}: Finally, multipliers $\mathcal{Q}_{l_{1}l_{2}}$ are updated as follows:
\begin{eqnarray}
\mathcal{Q}_{l_{1}l_{2}}^{+}=\mathcal{Q}_{l_{1}l_{2}}+\rho_{l_{1}l_{2}}(\mathcal{X}-\mathcal{Y}_{l_{1}l_{2}}).\label{UP2Q}
\end{eqnarray}
The optimization steps of the EMCP formulation are listed in Algorithm \ref{TC2}.
\begin{algorithm}[!h]
	\caption{EMCP} %算法的名字
	\hspace*{0.02in} {\bf Input:} %算法的输入,  \hspace*{0.02in}用来控制位置, 同时利用 \\ 进行换行
	An incomplete tensor $\mathcal{Z}$, the index set of the known elements $\Omega$, convergence criteria $\epsilon$, maximum iteration number $K$. \\
	\hspace*{0.02in} {\bf Initialization:} %算法的结果输出
	$\mathcal{X}^{0}=\mathcal{Z}_{\Omega}$, $\mathcal{Y}_{l_{1}l_{2}}^{0}=\mathcal{X}^{0}$, $\rho_{l_{1}l_{2}}^{0}>0$, $\mu>1$.
	\begin{algorithmic}
		%\State some description % \State 后写一般语句		
		\While{not converged and $k<K$} % While语句, 需要和EndWhile对应
		%\For{k=1,2,$\cdots$,K} % For 语句, 需要和EndFor对应	
		\State Updating $\mathcal{Y}_{l_{1}l_{2}}^{k}$ via (\ref{UP2Y});
		\State Updating $W_{l_{1}l_{2}}^{k}$ via (\ref{UP2W});
		\State Updating $\bar{\Lambda}_{l_{1}l_{2}}^{k}$ via (\ref{UP2LA});
		\State Updating $\mathcal{X}^{k}$ via (\ref{UP2X});
		\State Updating the multipliers $\mathcal{Q}_{l_{1}l_{2}}^{k}$ via (\ref{UP2Q});
		\State $\rho_{l_{1}l_{2}}^{k}=\mu\rho_{l_{1}l_{2}}^{k-1}$, $k=k+1$;
		\State Check the convergence conditions $\|\mathcal{X}^{k+1}-\mathcal{X}^{k}\|_{\infty}\leq\epsilon$.
		%	\EndFor
		\EndWhile
		\State \Return $\mathcal{X}^{k+1}$.
	\end{algorithmic}
	\hspace*{0.02in} {\bf Output:} %算法的结果输出
	Completed tensor $\mathcal{X}=\mathcal{X}^{k+1}$.
	\label{TC2}\end{algorithm}
\subsection{The BEMCP model}
Using the BEMCP Theorem, we get the following optimization model:
\begin{eqnarray}
\min_{\mathcal{X}}\sum_{1\leqslant l_{1}<l_{2}\leqslant N}^{N}\alpha_{l_{1}l_{2}}\|\mathcal{X}_{(l_{1}l_{2})}\|_{\Gamma_{l_{1}l_{2}},\Lambda_{l_{1}l_{2}}}\,s.t.\, \mathcal{P}_{\Omega}(\mathcal{X}-\mathcal{Z})=\mathbf{0}.\label{LRTCBEMCP}
\end{eqnarray}
Under the framework of the ADMM, the easy-to-implement optimization strategy could be provided to solve (\ref{LRTCBEMCP}). We introduce a set of tensors $\{\mathcal{Y}_{l_{1}l_{2}}=\mathcal{X}\}_{1\leqslant l_{1}<l_{2}\leqslant N}^{N}$ and transform optimization problem (\ref{LRTCBEMCP}), in its augmented Lagrangian form, as follows:
\begin{eqnarray}
&&J(\mathcal{X},\mathcal{Y}, \Lambda, \Gamma, \upsilon,\mathcal{Q})\nonumber
\\&&=\sum_{1\leqslant l_{1}<l_{2}\leqslant N}^{N}\alpha_{l_{1}l_{2}}\|\mathcal{Y}_{l_{1}l_{2}(l_{1}l_{2})}\|_{\Gamma_{l_{1}l_{2}},\Lambda_{l_{1}l_{2}}}\label{BEMCP2}
\\&&+\frac{\rho_{l_{1}l_{2}}}{2}\|\mathcal{X}-\mathcal{Y}_{l_{1}l_{2}}+\frac{\mathcal{Q}_{l_{1}l_{2}}}{\rho_{l_{1}l_{2}}}\|_{F}^{2} 
\quad s.t.\quad \mathcal{P}_{\Omega}(\mathcal{X}-\mathcal{Z})=\mathbf{0},\nonumber
\end{eqnarray}
where $\mathcal{Y}$ and $\mathcal{Q}$ are tensor sets; $\Lambda, \Gamma, \upsilon$ are matrix sets; $\{\mathcal{Y}_{l_{1}l_{2}}=\mathcal{X}\}_{1\leqslant l_{1}<l_{2}\leqslant N}^{N}$; $\{\mathcal{Q}_{l_{1}l_{2}}\}_{1\leqslant l_{1}<l_{2}\leqslant N}^{N}$ are Lagrangian multipliers; $\{\Lambda, \Gamma, \upsilon\}_{1\leqslant l_{1}<l_{2}\leqslant N}^{N}\in\mathbb{R}^{\mathit{I}_{3}\times R}$ are MCP variable sets; $\{\rho_{l_{1}l_{2}}\}_{1\leqslant l_{1}<l_{2}\leqslant N}^{N}>0$ are the augmented Lagrangian parameters; $\alpha_{l_{1}l_{2}}\geqslant0$ are weights and $\sum_{1\leqslant l_{1}<l_{2}\leqslant N}^{N}\alpha_{l_{1}l_{2}}=1$.

Besides, variables $\mathcal{X},\mathcal{Y}, \Lambda, \Gamma, \upsilon,\mathcal{Q}$ are updated alternately in the order of $\mathcal{Y}\to \upsilon\to \Lambda\to\Gamma\to\mathcal{X}\to\mathcal{Q}$. The update equations are derived in the following. 

\textbf{Update $\mathcal{Y}$}: Fix other variables, and the corresponding optimization are as follows: 
\begin{eqnarray}
\mathcal{Y}_{l_{1}l_{2}}^{+}&=&\arg\min_{\mathcal{Y}_{l_{1}l_{2}}}\alpha_{l_{1}l_{2}}\|\mathcal{Y}_{l_{1}l_{2}(l_{1}l_{2})}\|_{\Gamma_{l_{1}l_{2}},\Lambda_{l_{1}l_{2}}}\nonumber
\\&&+\frac{\rho_{l_{1}l_{2}}}{2}\|\mathcal{X}-\mathcal{Y}_{l_{1}l_{2}}+\frac{\mathcal{Q}_{l_{1}l_{2}}}{\rho_{l_{1}l_{2}}}\|_{F}^{2}. \nonumber
\end{eqnarray}
Calling Theorem \ref{thEWTGN}, the solution to the above optimization is given by:
\begin{eqnarray}
\mathcal{Y}_{l_{1}l_{2}}^{+}=\mathit{S}_{\frac{\Gamma_{l_{1}l_{2}}\rho_{l_{1}l_{2}}}{\alpha_{l_{1}l_{2}}},\frac{\Lambda_{l_{1}l_{2}}\rho_{l_{1}l_{2}}}{\alpha_{l_{1}l_{2}}}}(\mathcal{X}+\frac{\mathcal{Q}_{l_{1}l_{2}}}{\rho_{l_{1}l_{2}}}),\label{UP3Y}
\end{eqnarray}
where $\mathit{S}$ denotes the proximal operator defined in (\ref{opewtgn}).

\textbf{Update $\upsilon$}: Retaining only those components in $\mathcal{Y}_{l_{1}l_{2}}$ in ($\ref{BEMCP2}$) that depend on $\upsilon_{l_{1}l_{2}}$, we write
\begin{eqnarray}
\upsilon_{l_{1}l_{2}}^{+}=\arg\min_{\upsilon_{l_{1}l_{2}}}\|\mathcal{Y}_{l_{1}l_{2}(l_{1}l_{2})}\|_{\frac{\upsilon_{l_{1}l_{2}}}{\Gamma_{l_{1}l_{2}}},\ast}+\frac{1}{2}\|\frac{\upsilon_{l_{1}l_{2}}-\Lambda_{l_{1}l_{2}}\star\Gamma_{l_{1}l_{2}}}{\Gamma_{l_{1}l_{2}}\star\Gamma_{l_{1}l_{2}}}\|_{F}^{2},\nonumber
\end{eqnarray}
which has the following closed-form solution:
\begin{eqnarray}
\upsilon_{l_{1}l_{2}(i_{3},:)}^{+}=\max(\Lambda_{l_{1}l_{2}(i_{3},:)}\Gamma_{l_{1}l_{2}(i_{3},:)}-\sigma(\mathcal{Y}_{l_{1}l_{2}(l_{1}l_{2})}^{+(i_{3})}),\bar{\epsilon}),\label{UP3U}
\end{eqnarray}
where the element values of vector $\bar{\epsilon}$ are all small values close to 0, which will avoid the situation where $\Lambda$ becomes 0 and $\Gamma$ cannot be solved.

\textbf{Update $\Lambda$}: The update for $\Lambda_{l_{1}l_{2}}$ has the following closed-form solution:
\begin{eqnarray}
\Lambda_{l_{1}l_{2}}^{+}=\arg\min_{\Lambda_{l_{1}l_{2}}}\|\frac{\upsilon_{l_{1}l_{2}}-\Lambda_{l_{1}l_{2}}\star\Gamma_{l_{1}l_{2}}}{\Gamma_{l_{1}l_{2}}\star\Gamma_{l_{1}l_{2}}}\|_{F}^{2}=\frac{\upsilon_{l_{1}l_{2}}^{+}}{\Gamma_{l_{1}l_{2}}}.\label{UP3LA}
\end{eqnarray}

\textbf{Update $\Gamma$}: The update for $\Gamma_{l_{1}l_{2}}$ has the following closed-form solution:
\begin{eqnarray}
\Gamma_{l_{1}l_{2}}^{+}=\arg\min_{\Gamma_{l_{1}l_{2}}}\|\mathcal{Y}^{+}_{l_{1}l_{2}}\|_{\frac{\upsilon^{+}_{l_{1}l_{2}}}{\Gamma_{l_{1}l_{2}}},\ast}+\frac{1}{2}\|\frac{\upsilon^{+}_{l_{1}l_{2}}-\Lambda^{+}_{l_{1}l_{2}}\star\Gamma_{l_{1}l_{2}}}{\Gamma_{l_{1}l_{2}}\star\Gamma_{l_{1}l_{2}}}\|_{F}^{2}.\label{UPGA}
\end{eqnarray}
Problem (\ref{UPGA}) is element-wise separable, and by proposition \ref{proposition1} we have the following results:
\begin{eqnarray}
&&\|\mathcal{Y}\|_{\Gamma_{l_{1}l_{2}},\Lambda_{l_{1}l_{2}}}\nonumber
\\&=&\sum_{i=1}^{\mathit{I}_{3}}\sum_{j=1}^{R}\frac{2\upsilon_{(i,j)}\sigma_{j}(\bar{\mathcal{Y}}^{(i)})+(\upsilon_{(i,j)}-\Lambda_{(i,j)}\Gamma_{(i,j)})^{2}}{2\Gamma_{(i,j)}}.\nonumber
\end{eqnarray}
Then, we consider the case of one of the elements individually:
\begin{eqnarray}
\frac{2\upsilon_{(i,j)}\sigma_{j}(\bar{\mathcal{Y}}^{(i)})+(\upsilon_{(i,j)}-\Lambda_{(i,j)}\Gamma_{(i,j)})^{2}}{2\Gamma_{(i,j)}}\label{GA2}.
\end{eqnarray}
The closed form of $\Gamma_{(i,j)}$ can be derived by setting the derivative of (\ref{GA2}) to zero:
\begin{eqnarray}
\frac{(\Lambda_{(i,j)}\Gamma_{(i,j)})^{2}-2\upsilon_{(i,j)}\sigma_{j}(\bar{\mathcal{Y}}^{(i)})-\upsilon_{(i,j)}^{2}}{2\Gamma_{(i,j)}^{2}}=0.\nonumber
\end{eqnarray}
So, $\Gamma_{(i,j)}$ is updated by the following:
\begin{eqnarray}
\Gamma^{+}_{(i,j)}=\sqrt{\frac{2\upsilon^{+}_{(i,j)}\sigma_{j}(\bar{\mathcal{Y}}^{+(i)})+\upsilon_{(i,j)}^{+2}}{\Lambda_{(i,j)}^{+2}}}.\label{UP3GA}
\end{eqnarray}

\textbf{Update $\mathcal{X}$}: The closed form of $\mathcal{X}$ can be derived by setting the derivative of (\ref{BEMCP2}) to zero. We can now update $\mathcal{X}$ by the following equation:
\begin{eqnarray}
\mathcal{X}^{+}=\mathcal{P}_{\Omega^{c}}(\frac{\sum_{1\leqslant l_{1}< l_{2}\leqslant N}\rho_{l_{1}l_{2}}(\mathcal{Y}_{l_{1}l_{2}}-\frac{\mathcal{Q}_{l_{1}l_{2}}}{\rho_{l_{1}l_{2}}})}{\sum_{1\leqslant l_{1}< l_{2}\leqslant N}\rho_{l_{1}l_{2}}})+\mathcal{P}_{\Omega}(\mathcal{Z}).\label{UP3X}
\end{eqnarray}

\textbf{Update $\mathcal{Q}$}: Finally, multipliers $\mathcal{Q}_{l_{1}l_{2}}$ are updated as follows:
\begin{eqnarray}
\mathcal{Q}_{l_{1}l_{2}}^{+}=\mathcal{Q}_{l_{1}l_{2}}+\rho_{l_{1}l_{2}}(\mathcal{X}-\mathcal{Y}_{l_{1}l_{2}}).\label{UP3Q}
\end{eqnarray}
The optimization steps of BEMCP formulation are listed in Algorithm \ref{TC3}. The main per-iteration cost lies in the update of $\mathcal{Y}_{l_{1}l_{2}}$, which requires computing t-SVD. The per-iteration complexity is $O(LE(\sum_{1\leqslant l_{1}< l_{2}\leqslant N}[log(le_{l_{1}l_{2}})+\min(\mathit{I}_{l_{1}},\mathit{I}_{l_{2}})]))$, where $LE=\prod_{i=1}^{N}\mathit{I}_{i}$ and $le_{l_{1}l_{2}}=LE/(\mathit{I}_{l_{1}}\mathit{I}_{l_{2}})$.
\begin{algorithm}[!h]
	\caption{BEMCP} %算法的名字
	\hspace*{0.02in} {\bf Input:} %算法的输入,  \hspace*{0.02in}用来控制位置, 同时利用 \\ 进行换行
	An incomplete tensor $\mathcal{Z}$, the index set of the known elements $\Omega$, convergence criteria $\epsilon$, maximum iteration number $K$. \\
	\hspace*{0.02in} {\bf Initialization:} %算法的结果输出
	$\mathcal{X}^{0}=\mathcal{Z}_{\Omega}$, $\mathcal{Y}_{l_{1}l_{2}}^{0}=\mathcal{X}^{0}$, $\rho_{l_{1}l_{2}}^{0}>0$, $\mu>1$.
	\begin{algorithmic}
		%\State some description % \State 后写一般语句		
		\While{not converged and $k<K$} % While语句, 需要和EndWhile对应
		%\For{k=1,2,$\cdots$,K} % For 语句, 需要和EndFor对应	
		\State Updating $\mathcal{Y}_{l_{1}l_{2}}^{k}$ via (\ref{UP3Y});
		\State Updating $W_{l_{1}l_{2}}^{k}$ via (\ref{UP3U});
		\State Updating $\Lambda_{l_{1}l_{2}}^{k}$ via (\ref{UP3LA});
		\State Updating $\Gamma_{l_{1}l_{2}}^{k}$ via (\ref{UP3GA});
		\State Updating $\mathcal{X}^{k}$ via (\ref{UP3X});
		\State Updating the multipliers $\mathcal{Q}_{l_{1}l_{2}}^{k}$ via (\ref{UP3Q});
		\State $\rho_{l_{1}l_{2}}^{k}=\mu\rho_{l_{1}l_{2}}^{k-1}$, $k=k+1$;
		\State Check the convergence conditions $\|\mathcal{X}^{k+1}-\mathcal{X}^{k}\|_{\infty}\leq\epsilon$.
		%	\EndFor
		\EndWhile
		\State \Return $\mathcal{X}^{k+1}$.
	\end{algorithmic}
	\hspace*{0.02in} {\bf Output:} %算法的结果输出
	Completed tensor $\mathcal{X}=\mathcal{X}^{k+1}$.
	\label{TC3}\end{algorithm}

\section{EXPERIMENTS}
We evaluate the performance of the proposed LRTC methods. We employ the peak signal-to-noise rate (PSNR) value, the structural similarity (SSIM) value \cite{1284395}, the feature similarity (FSIM) value \cite{5705575}, and erreur relative globale adimensionnelle de synth$\grave{e}$se (ERGAS) value \cite{2432002352} to measure the quality of the recovered results. The PSNR, SSIM, and FSIM values are the bigger the better, and the ERGAS value is the smaller the better. All tests are implemented on the Windows 10 platform and MATLAB (R2019a) with an Intel Core i7-10875H 2.30 GHz and 32 GB of RAM.

In this section, we test three kinds of real-world data: MSI, MRI, and CV. The method for sampling the data is purely random sampling. The comparative LRTC methods are as follows: HaLRTC \cite{6138863}, LRTCTV-I \cite{3120171} represent state-of-the-art for the Tucker-decomposition-based methods; and TNN \cite{7782758}, PSTNN \cite{1122020112680}, FTNN \cite{9115254}, WSTNN \cite{2020170} represent state-of-the-art for the t-SVD-based methods. Since the TNN, PSTNN, and FTNN methods only apply to three-order tensors, in all four-order tensor tests, we reshape them into three-order tensors and then test the performance of these methods.

\subsection{MSI completion}

\begin{figure*}[!h] %这里使用的是强制位置, 除非真的放不下, 不然就是写在哪里图就放在哪里, 不会乱动
	\centering  %图片全局居中
	\vspace{0cm} %设置与上面正文的距离
	%	\subfigtopskip=2pt %设置子图与上面正文或别的内容的距离
	%		\subfigbottomskip=2pt %设置第二行子图与第一行子图的距离, 即下面的头与上面的脚的距离ttt
	%	\subfigcapskip=-5pt %设置子图与子标题之间的距离
	\subfloat[]{
		\begin{minipage}[b]{0.076\linewidth}
			\includegraphics[width=1\linewidth]{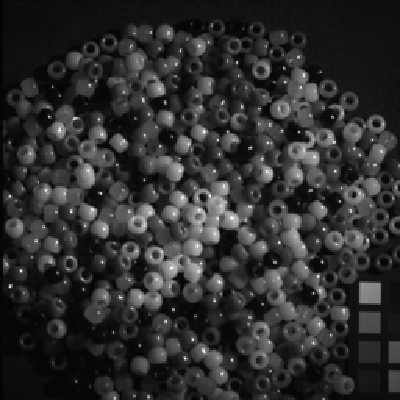}\\
			
			\includegraphics[width=1\linewidth]{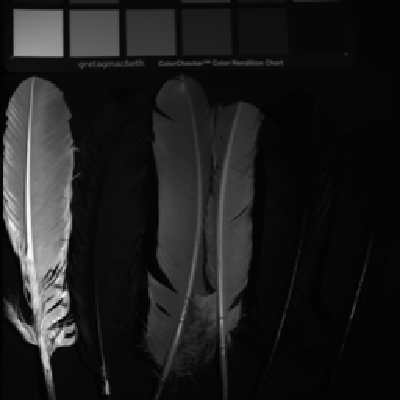}\\
			
			\includegraphics[width=1\linewidth]{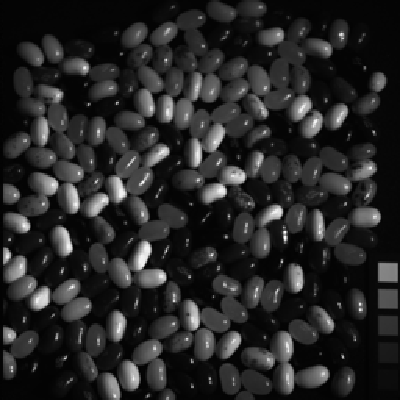}
	\end{minipage}}
	\subfloat[]{
		\begin{minipage}[b]{0.076\linewidth}
			\includegraphics[width=1\linewidth]{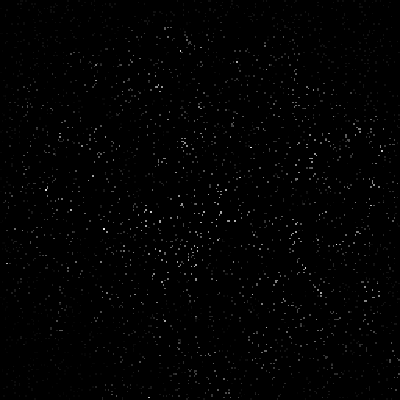}\\
			
			\includegraphics[width=1\linewidth]{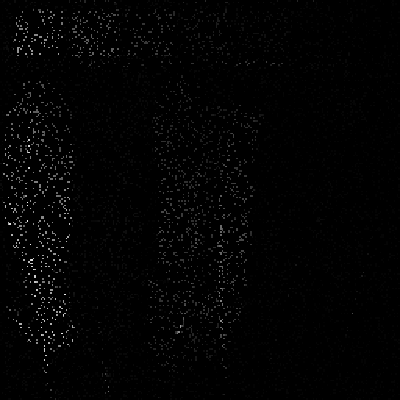}\\
			
			\includegraphics[width=1\linewidth]{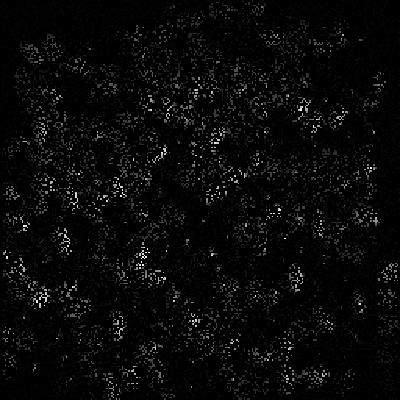}
	\end{minipage}}
	\subfloat[]{
		\begin{minipage}[b]{0.076\linewidth}
			\includegraphics[width=1\linewidth]{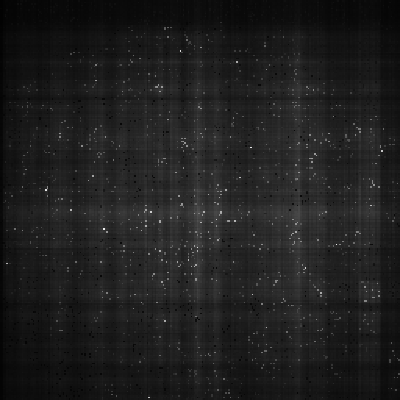}\\
			
			\includegraphics[width=1\linewidth]{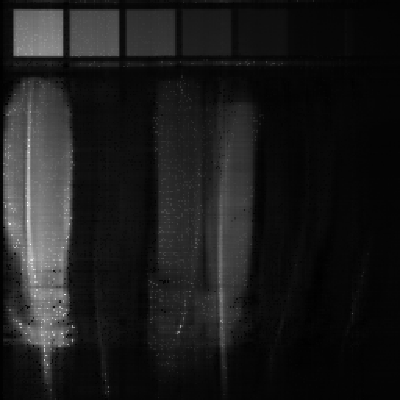}\\
			
			\includegraphics[width=1\linewidth]{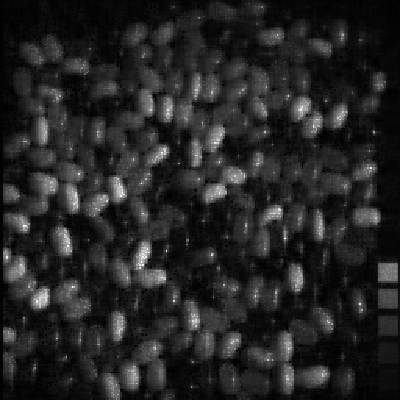}
	\end{minipage}}
	\subfloat[]{
		\begin{minipage}[b]{0.076\linewidth}
			\includegraphics[width=1\linewidth]{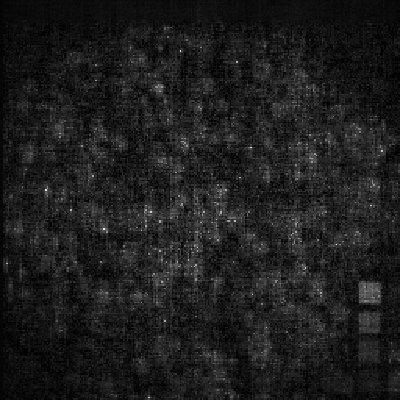}\\
			
			\includegraphics[width=1\linewidth]{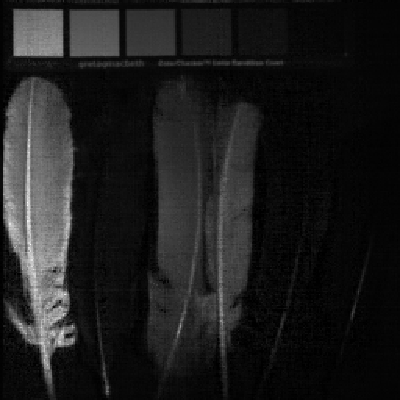}\\
			
			\includegraphics[width=1\linewidth]{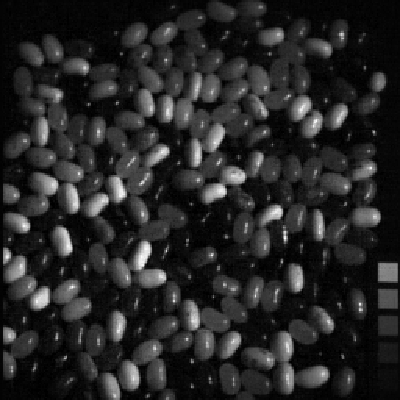}
	\end{minipage}}
	\subfloat[]{
		\begin{minipage}[b]{0.076\linewidth}
			\includegraphics[width=1\linewidth]{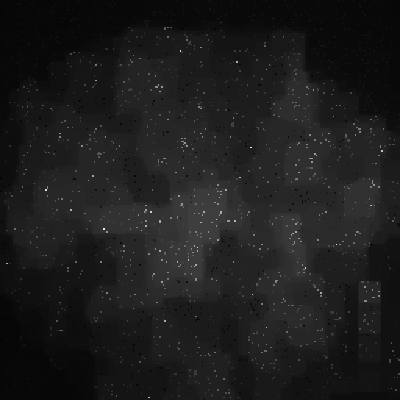}\\
			
			\includegraphics[width=1\linewidth]{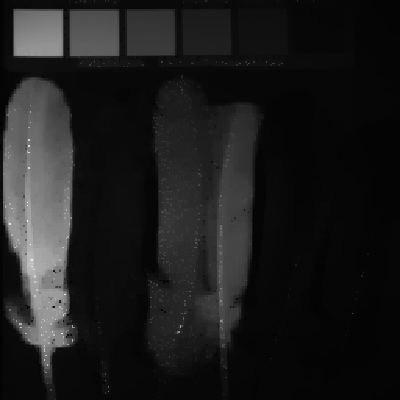}\\
			
			\includegraphics[width=1\linewidth]{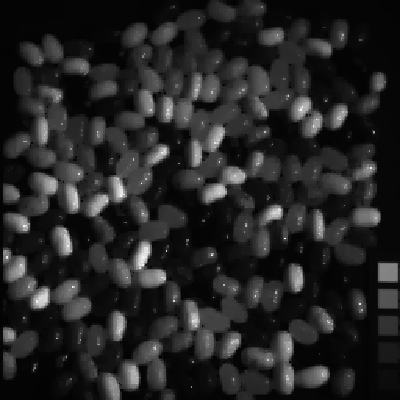}
	\end{minipage}}
	\subfloat[]{
		\begin{minipage}[b]{0.076\linewidth}
			\includegraphics[width=1\linewidth]{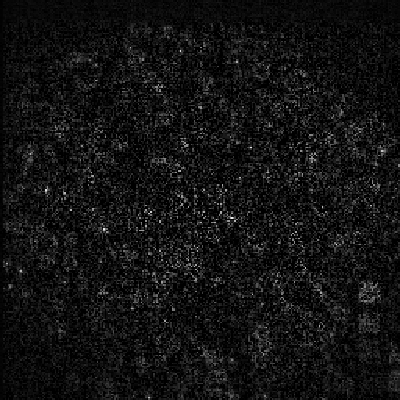}\\

			\includegraphics[width=1\linewidth]{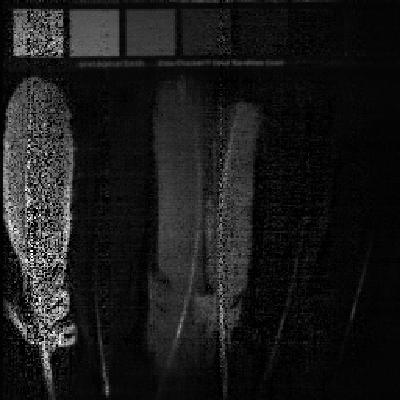}\\

			\includegraphics[width=1\linewidth]{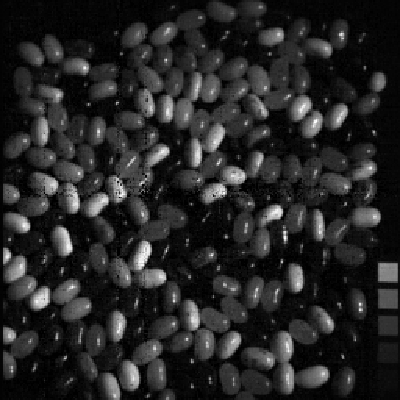}
	\end{minipage}}
	\subfloat[]{
		\begin{minipage}[b]{0.076\linewidth}
			\includegraphics[width=1\linewidth]{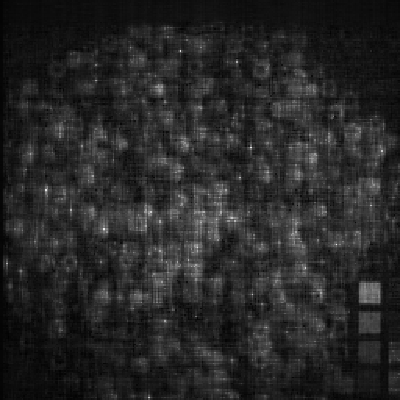}\\

			\includegraphics[width=1\linewidth]{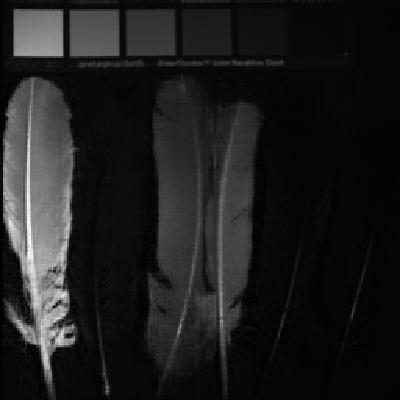}\\

			\includegraphics[width=1\linewidth]{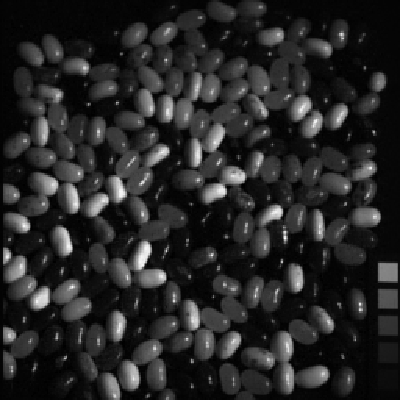}
	\end{minipage}}
	\subfloat[]{
		\begin{minipage}[b]{0.076\linewidth}
			\includegraphics[width=1\linewidth]{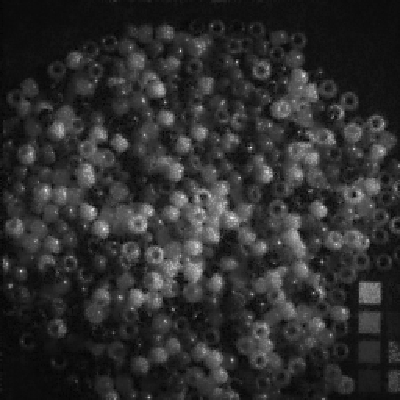}\\

			\includegraphics[width=1\linewidth]{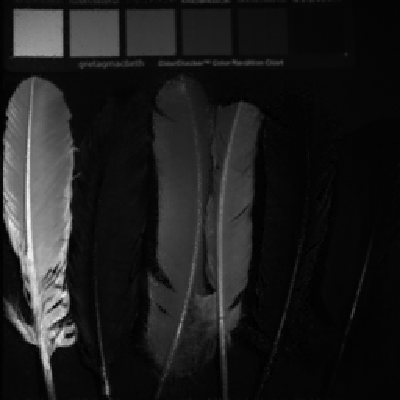}\\

			\includegraphics[width=1\linewidth]{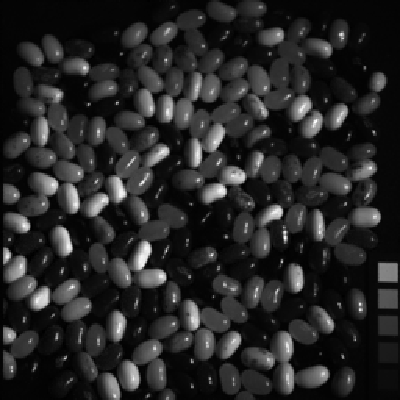}
	\end{minipage}}
	\subfloat[]{
		\begin{minipage}[b]{0.076\linewidth}
			\includegraphics[width=1\linewidth]{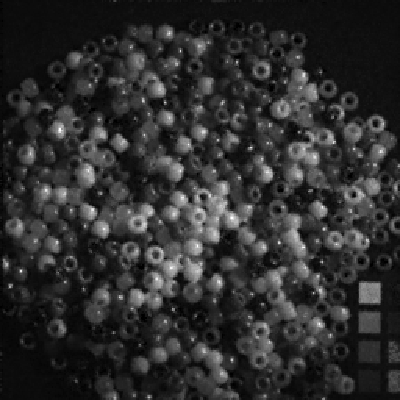}\\

			\includegraphics[width=1\linewidth]{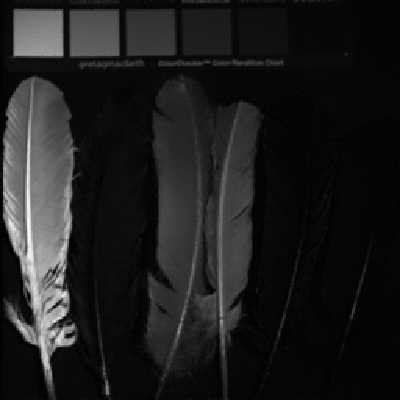}\\

			\includegraphics[width=1\linewidth]{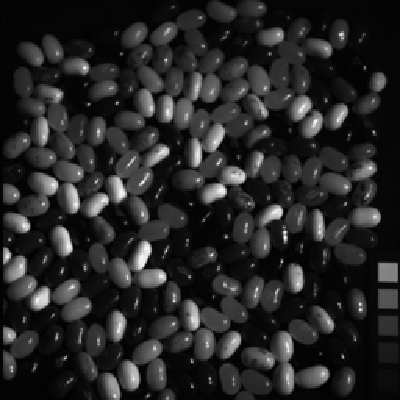}
	\end{minipage}}
	\subfloat[]{
		\begin{minipage}[b]{0.076\linewidth}
			\includegraphics[width=1\linewidth]{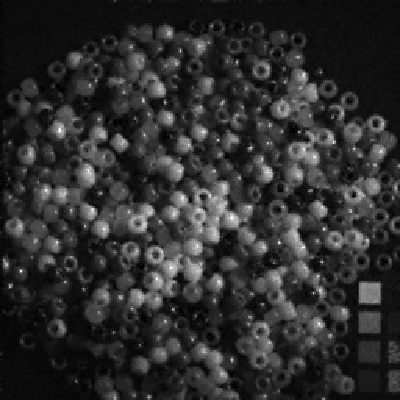}\\

			\includegraphics[width=1\linewidth]{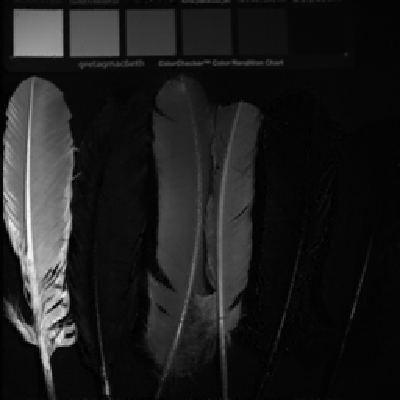}\\

			\includegraphics[width=1\linewidth]{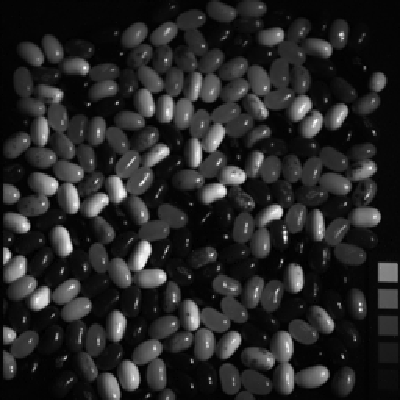}
	\end{minipage}}
	\subfloat[]{
		\begin{minipage}[b]{0.076\linewidth}
			\includegraphics[width=1\linewidth]{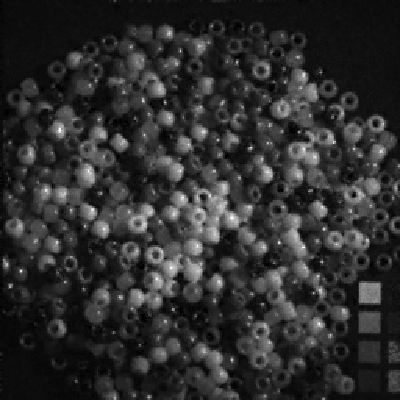}\\

			\includegraphics[width=1\linewidth]{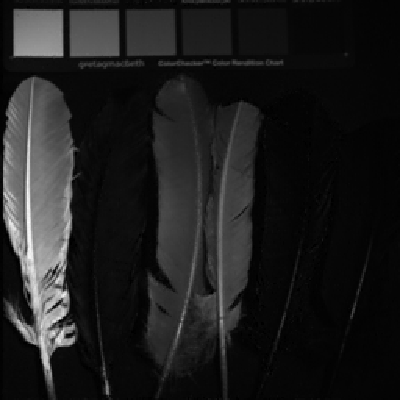}\\
			
			\includegraphics[width=1\linewidth]{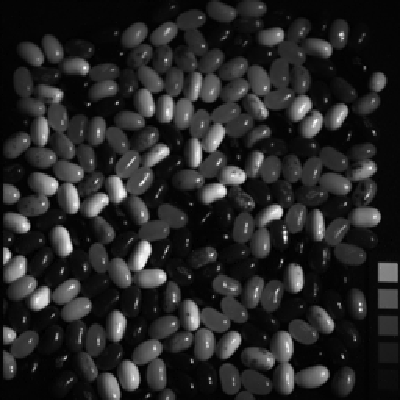}
	\end{minipage}}
	\caption{(a) Original image. (b) Observed image. (c) HaLRTC. (d) TNN. (e) LRTCTV-I. (f) PSTNN. (g) FTNN. (h) WSTNN. (i) NMCP. (j) EMCP. (k) BEMCP.  SR: top row is 5\%, middle row is 10\% and last row is 20\%. The rows of MSIs are in order: beads, feathers, jelly\_beans. The corresponding bands in each row are: 31, 12, 20.}
	\label{MSITC}
\end{figure*}

We test 32 MSIs in the dataset CAVE\footnote{http://www.cs.columbia.edu/CAVE/databases/multispectral/}. All testing data are of size $256\times256\times31$. In Fig.\ref{MSITC}, we randomly select three from 32 MSIs, bringing the different sampling rates and band visible results. The individual MSI names and their corresponding bands are written in the caption of Fig.\ref{MSITC}. As shown from Fig.\ref{MSITC}, the visual effect of the BEMCP is better than the contrast method at all three sampling rates. To further highlight the superiority of our method, the average quantitative results of 32 MSIs are listed in Table \ref{MSITC1}. It can be seen that the three mothods proposed in this paper have a great improvement compared to the WSTNN method. The PSNR value at both 10\% and 20\% sampling rate is at least 1.5dB higher than the WSTNN method, and even reaches 5dB at 5\% sampling rate. Besides, the results show that the PSNR value of the BEMCP method is 0.2db higher than that of the EMCP method at all three sampling rates. This indicates that the BEMCP method is better than the univariate EMCP method. And compared with the NMCP method that directly uses the MCP function, our improvement is more prominent. More experimental results are available in Appendix A.
\begin{table*}[]
	\caption{The average PSNR, SSIM, FSIM and ERGAS values for 32 MSIs tested by observed and the nine utilized LRTC methods.}
	\resizebox{\textwidth}{!}{
		\begin{tabular}{|c|cccc|cccc|cccc|c|}
			\hline
			SR             & \multicolumn{4}{c|}{5\%}                                                                                                           & \multicolumn{4}{c|}{10\%}                                                                                                          & \multicolumn{4}{c|}{20\%}                                                                                                          & Time(s) \\ \hline
			Method         & \multicolumn{1}{c|}{PSNR}            & \multicolumn{1}{c|}{SSIM}           & \multicolumn{1}{c|}{FSIM}           & ERGAS           & \multicolumn{1}{c|}{PSNR}            & \multicolumn{1}{c|}{SSIM}           & \multicolumn{1}{c|}{FSIM}           & ERGAS           & \multicolumn{1}{c|}{PSNR}            & \multicolumn{1}{c|}{SSIM}           & \multicolumn{1}{c|}{FSIM}           & ERGAS           &         \\ \hline
			Observed       & \multicolumn{1}{c|}{15.438}          & \multicolumn{1}{c|}{0.153}          & \multicolumn{1}{c|}{0.644}          & 845.339         & \multicolumn{1}{c|}{15.673}          & \multicolumn{1}{c|}{0.194}          & \multicolumn{1}{c|}{0.646}          & 822.808         & \multicolumn{1}{c|}{16.185}          & \multicolumn{1}{c|}{0.269}          & \multicolumn{1}{c|}{0.651}          & 775.716         & 0.000   \\ \hline
			HaLRTC         & \multicolumn{1}{c|}{25.367}          & \multicolumn{1}{c|}{0.774}          & \multicolumn{1}{c|}{0.837}          & 298.654         & \multicolumn{1}{c|}{29.855}          & \multicolumn{1}{c|}{0.856}          & \multicolumn{1}{c|}{0.894}          & 184.887         & \multicolumn{1}{c|}{35.038}          & \multicolumn{1}{c|}{0.930}          & \multicolumn{1}{c|}{0.946}          & 105.307         & 26.656  \\ \hline
			TNN            & \multicolumn{1}{c|}{25.350}          & \multicolumn{1}{c|}{0.713}          & \multicolumn{1}{c|}{0.817}          & 289.617         & \multicolumn{1}{c|}{33.114}          & \multicolumn{1}{c|}{0.880}          & \multicolumn{1}{c|}{0.918}          & 127.987         & \multicolumn{1}{c|}{40.201}          & \multicolumn{1}{c|}{0.964}          & \multicolumn{1}{c|}{0.972}          & 58.856          & 85.578  \\ \hline
			LRTCTV-I       & \multicolumn{1}{c|}{25.886}          & \multicolumn{1}{c|}{0.800}          & \multicolumn{1}{c|}{0.835}          & 276.943         & \multicolumn{1}{c|}{30.725}          & \multicolumn{1}{c|}{0.890}          & \multicolumn{1}{c|}{0.906}          & 162.443         & \multicolumn{1}{c|}{35.516}          & \multicolumn{1}{c|}{0.949}          & \multicolumn{1}{c|}{0.957}          & 94.262          & 449.471 \\ \hline
			PSTNN          & \multicolumn{1}{c|}{18.708}          & \multicolumn{1}{c|}{0.474}          & \multicolumn{1}{c|}{0.650}          & 574.923         & \multicolumn{1}{c|}{23.211}          & \multicolumn{1}{c|}{0.683}          & \multicolumn{1}{c|}{0.782}          & 352.958         & \multicolumn{1}{c|}{34.315}          & \multicolumn{1}{c|}{0.924}          & \multicolumn{1}{c|}{0.942}          & 116.434         & 91.420  \\ \hline
			FTNN           & \multicolumn{1}{c|}{32.645}          & \multicolumn{1}{c|}{0.899}          & \multicolumn{1}{c|}{0.924}          & 131.419         & \multicolumn{1}{c|}{37.151}          & \multicolumn{1}{c|}{0.954}          & \multicolumn{1}{c|}{0.963}          & 78.977          & \multicolumn{1}{c|}{43.023}          & \multicolumn{1}{c|}{0.984}          & \multicolumn{1}{c|}{0.987}          & 41.714          & 503.684 \\ \hline
			WSTNN          & \multicolumn{1}{c|}{31.431}          & \multicolumn{1}{c|}{0.806}          & \multicolumn{1}{c|}{0.911}          & 208.954         & \multicolumn{1}{c|}{40.143}          & \multicolumn{1}{c|}{0.981}          & \multicolumn{1}{c|}{0.981}          & 53.010          & \multicolumn{1}{c|}{47.049}          & \multicolumn{1}{c|}{0.995}          & \multicolumn{1}{c|}{0.995}          & 24.974          & 125.812 \\ \hline
			NMCP           & \multicolumn{1}{c|}{37.474}          & \multicolumn{1}{c|}{0.962}          & \multicolumn{1}{c|}{0.962}          & 70.020          & \multicolumn{1}{c|}{42.673}          & \multicolumn{1}{c|}{0.987}          & \multicolumn{1}{c|}{0.987}          & 39.156          & \multicolumn{1}{c|}{48.599}          & \multicolumn{1}{c|}{0.996}          & \multicolumn{1}{c|}{0.996}          & 20.403          & 177.065 \\ \hline
			EMCP           & \multicolumn{1}{c|}{37.602}          & \multicolumn{1}{c|}{0.960}          & \multicolumn{1}{c|}{0.961}          & 69.286          & \multicolumn{1}{c|}{43.507}          & \multicolumn{1}{c|}{0.987}          & \multicolumn{1}{c|}{0.987}          & 35.883          & \multicolumn{1}{c|}{49.668}          & \multicolumn{1}{c|}{0.995}          & \multicolumn{1}{c|}{0.996}          & 17.996          & 194.907 \\ \hline
			BEMCP & \multicolumn{1}{c|}{\textbf{37.939}} & \multicolumn{1}{c|}{\textbf{0.963}} & \multicolumn{1}{c|}{\textbf{0.963}} & \textbf{66.775} & \multicolumn{1}{c|}{\textbf{43.734}} & \multicolumn{1}{c|}{\textbf{0.988}} & \multicolumn{1}{c|}{\textbf{0.988}} & \textbf{34.998} & \multicolumn{1}{c|}{\textbf{49.904}} & \multicolumn{1}{c|}{\textbf{0.996}} & \multicolumn{1}{c|}{\textbf{0.996}} & \textbf{17.551} & 204.299 \\ \hline
	\end{tabular}}\label{MSITC1}
\end{table*}

\subsection{MRI completion}

\begin{figure*}[!h] %这里使用的是强制位置, 除非真的放不下, 不然就是写在哪里图就放在哪里, 不会乱动
	\centering  %图片全局居中
	\vspace{0cm} %设置与上面正文的距离
	%	\subfigtopskip=2pt %设置子图与上面正文或别的内容的距离
	%	\subfigbottomskip=2pt %设置第二行子图与第一行子图的距离, 即下面的头与上面的脚的距离
	%	\subfigcapskip=-5pt %设置子图与子标题之间的距离
	\subfloat[]{
		\begin{minipage}[b]{0.076\linewidth}
			\includegraphics[width=1\linewidth]{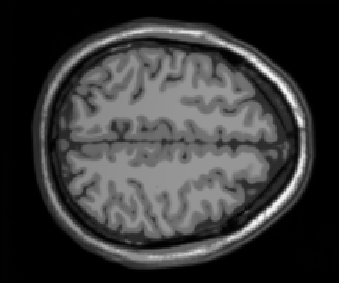}\\
			
			\includegraphics[width=1\linewidth]{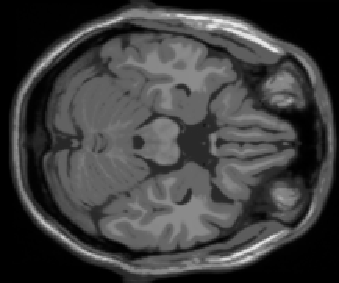}\\
			
			\includegraphics[width=1\linewidth]{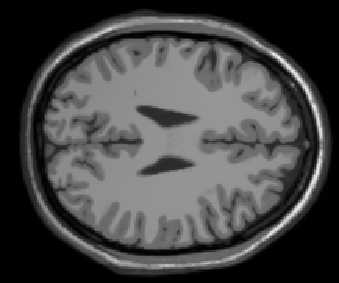}
	\end{minipage}}
	\subfloat[]{
		\begin{minipage}[b]{0.076\linewidth}
			\includegraphics[width=1\linewidth]{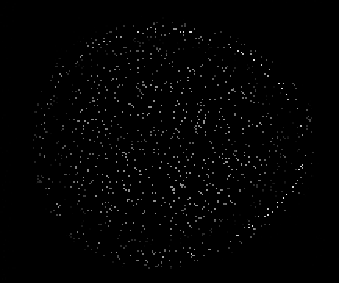}\\
			
			\includegraphics[width=1\linewidth]{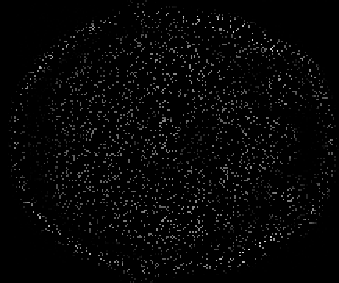}\\
			
			\includegraphics[width=1\linewidth]{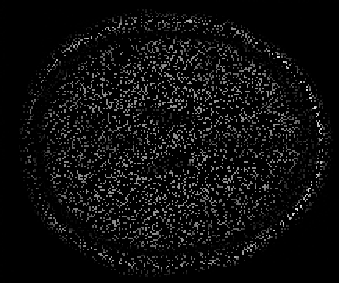}
	\end{minipage}}
	\subfloat[]{
		\begin{minipage}[b]{0.076\linewidth}
			\includegraphics[width=1\linewidth]{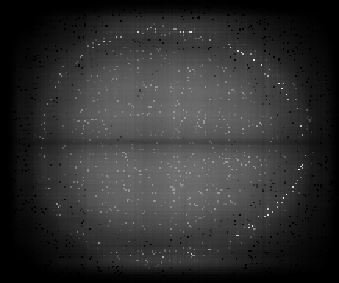}\\
			
			\includegraphics[width=1\linewidth]{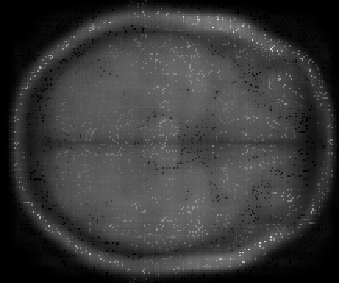}\\
			
			\includegraphics[width=1\linewidth]{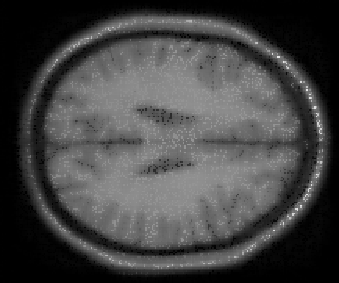}
	\end{minipage}}
	\subfloat[]{
		\begin{minipage}[b]{0.076\linewidth}
			\includegraphics[width=1\linewidth]{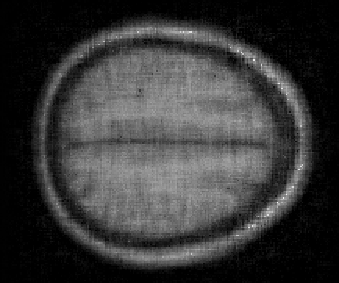}\\
			
			\includegraphics[width=1\linewidth]{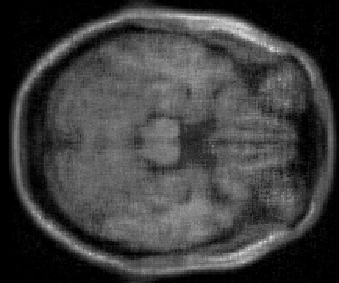}\\
			
			\includegraphics[width=1\linewidth]{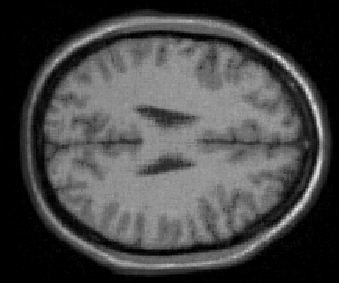}
	\end{minipage}}
	\subfloat[]{
		\begin{minipage}[b]{0.076\linewidth}
			\includegraphics[width=1\linewidth]{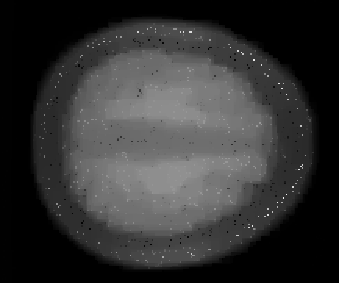}\\
			
			\includegraphics[width=1\linewidth]{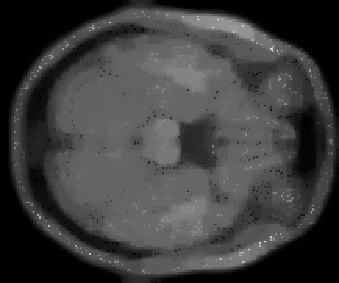}\\
			
			\includegraphics[width=1\linewidth]{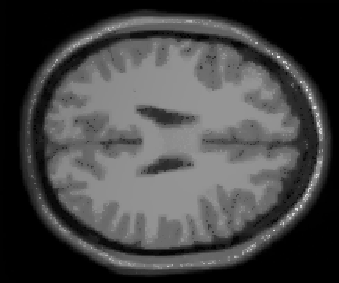}
	\end{minipage}}
	\subfloat[]{
		\begin{minipage}[b]{0.076\linewidth}
			\includegraphics[width=1\linewidth]{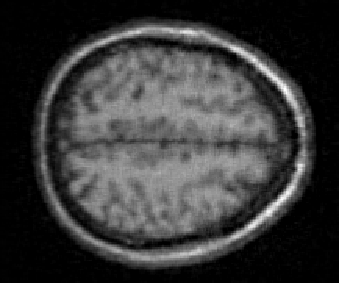}\\
			
			\includegraphics[width=1\linewidth]{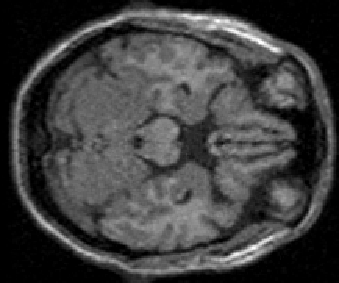}\\
			
			\includegraphics[width=1\linewidth]{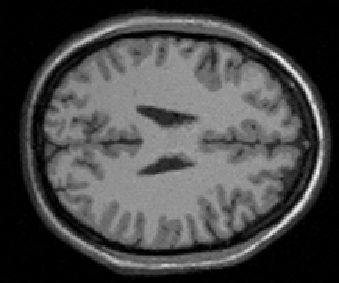}
	\end{minipage}}
	\subfloat[]{
		\begin{minipage}[b]{0.076\linewidth}
			\includegraphics[width=1\linewidth]{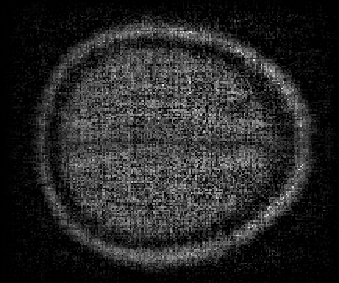}\\
			
			\includegraphics[width=1\linewidth]{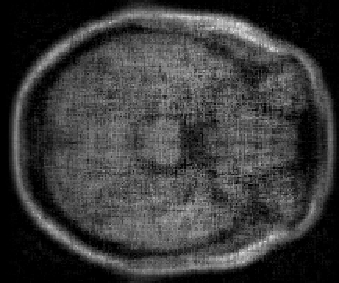}\\
			
			\includegraphics[width=1\linewidth]{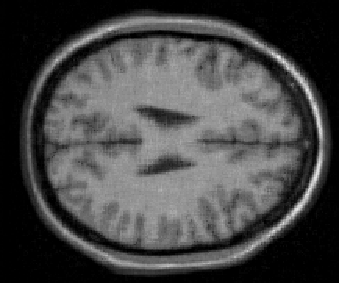}
	\end{minipage}}
	\subfloat[]{
		\begin{minipage}[b]{0.076\linewidth}
			\includegraphics[width=1\linewidth]{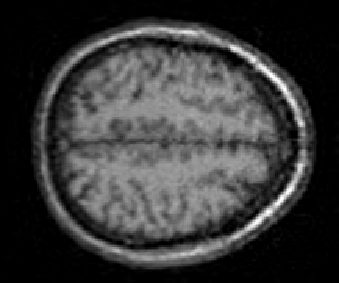}\\
			
			\includegraphics[width=1\linewidth]{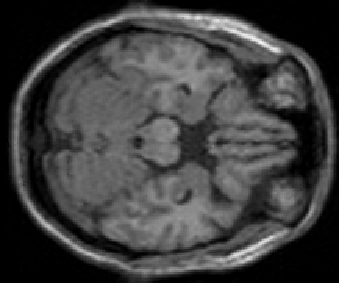}\\
			
			\includegraphics[width=1\linewidth]{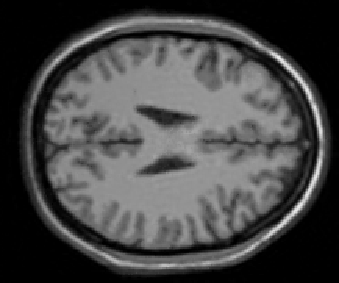}
	\end{minipage}}
	\subfloat[]{
		\begin{minipage}[b]{0.076\linewidth}
			\includegraphics[width=1\linewidth]{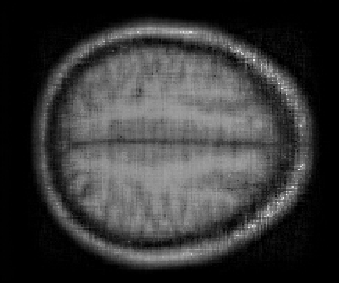}\\
			
			\includegraphics[width=1\linewidth]{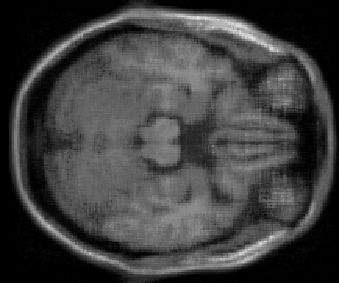}\\
			
			\includegraphics[width=1\linewidth]{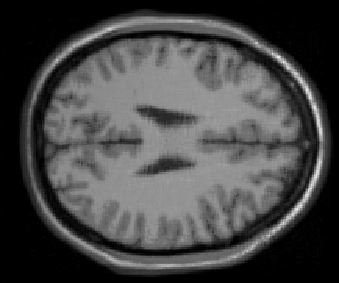}
	\end{minipage}}
	\subfloat[]{
		\begin{minipage}[b]{0.076\linewidth}
			\includegraphics[width=1\linewidth]{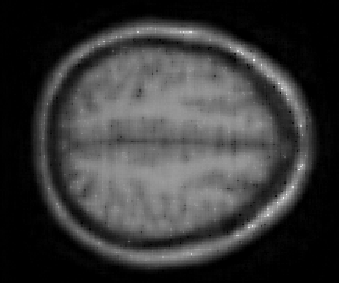}\\
			
			\includegraphics[width=1\linewidth]{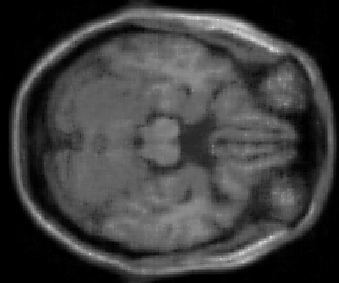}\\
			
			\includegraphics[width=1\linewidth]{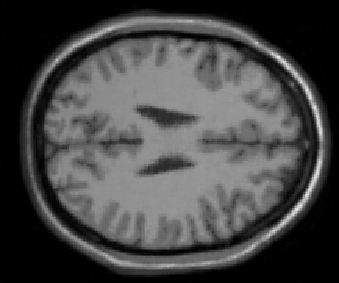}
	\end{minipage}}
	\subfloat[]{
		\begin{minipage}[b]{0.076\linewidth}
			\includegraphics[width=1\linewidth]{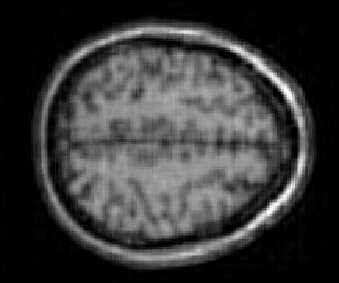}\\
			
			\includegraphics[width=1\linewidth]{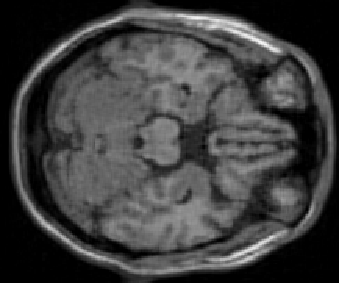}\\
			
			\includegraphics[width=1\linewidth]{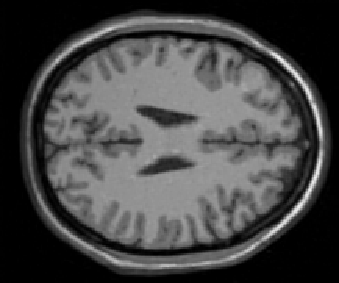}
	\end{minipage}}
	\caption{(a) Original image. (b) Observed image. (c) HaLRTC. (d) TNN. (e) LRTCTV-I. (f) PSTNN. (g) FTNN. (h) WSTNN. (i) NMCP. (j) EMCP. (k) BEMCP.  Each type of slice: the first row is the 120th slice with a sampling rate of 5\%, the second row is the 50th slice with a sampling rate of 10\%, and the third row is the 100th slice with a sampling rate of 20\%.}
	\label{MRITC}
\end{figure*}

We test the performance of the proposed method and the comparative method on MRI\footnote{http://brainweb.bic.mni.mcgill.ca/brainweb/selection\_normal.html} data with the size of $181\times217\times181$. First, we demonstrate the visual effect recovered by MRI data at sampling rates of 5\%, 10\% and 20\% in Fig.\ref{MRITC}. Our method is clearly superior to the comparative methods. Then, we list the average quantitative results of frontal sections of MRI restored by all methods at different sampling rates in Table \ref{MRITC1}. At the sampling rate of 5\% and 10\%, the PSNR value of the three methods is at least 1db higher than that of the WSTNN method. The PSNR value obtained by the BEMCP method is 0.4dB higher than that of the EMCP method at the sampling rate of 5\% and 10\%, and the values of SSIM, FSIM, and ERGAS are also better than the EMCP method.
\begin{table*}[]
	\caption{The PSNR, SSIM, FSIM and ERGAS values output by by observed and the nine utilized LRTC methods for MRI.}
	\resizebox{\textwidth}{!}{
		\begin{tabular}{|c|cccc|cccc|cccc|c|}
			\hline
			SR             & \multicolumn{4}{c|}{5\%}                                                                                                            & \multicolumn{4}{c|}{10\%}                                                                                                          & \multicolumn{4}{c|}{20\%}                                                                                                          & Time(s)  \\ \hline
			Method         & \multicolumn{1}{c|}{PSNR}            & \multicolumn{1}{c|}{SSIM}           & \multicolumn{1}{c|}{FSIM}           & ERGAS            & \multicolumn{1}{c|}{PSNR}            & \multicolumn{1}{c|}{SSIM}           & \multicolumn{1}{c|}{FSIM}           & ERGAS           & \multicolumn{1}{c|}{PSNR}            & \multicolumn{1}{c|}{SSIM}           & \multicolumn{1}{c|}{FSIM}           & ERGAS           &          \\ \hline
			Observed       & \multicolumn{1}{c|}{11.399}          & \multicolumn{1}{c|}{0.310}          & \multicolumn{1}{c|}{0.530}          & 1021.050         & \multicolumn{1}{c|}{11.634}          & \multicolumn{1}{c|}{0.323}          & \multicolumn{1}{c|}{0.565}          & 993.910         & \multicolumn{1}{c|}{12.147}          & \multicolumn{1}{c|}{0.350}          & \multicolumn{1}{c|}{0.613}          & 936.861         & 0.000    \\ \hline
			HaLRTC         & \multicolumn{1}{c|}{17.302}          & \multicolumn{1}{c|}{0.298}          & \multicolumn{1}{c|}{0.637}          & 537.239          & \multicolumn{1}{c|}{20.099}          & \multicolumn{1}{c|}{0.438}          & \multicolumn{1}{c|}{0.725}          & 391.488         & \multicolumn{1}{c|}{24.454}          & \multicolumn{1}{c|}{0.660}          & \multicolumn{1}{c|}{0.829}          & 235.314         & 62.281   \\ \hline
			TNN            & \multicolumn{1}{c|}{22.730}          & \multicolumn{1}{c|}{0.473}          & \multicolumn{1}{c|}{0.743}          & 301.839          & \multicolumn{1}{c|}{26.073}          & \multicolumn{1}{c|}{0.643}          & \multicolumn{1}{c|}{0.812}          & 205.892         & \multicolumn{1}{c|}{29.976}          & \multicolumn{1}{c|}{0.799}          & \multicolumn{1}{c|}{0.882}          & 130.784         & 266.988  \\ \hline
			LRTCTV-I       & \multicolumn{1}{c|}{19.369}          & \multicolumn{1}{c|}{0.597}          & \multicolumn{1}{c|}{0.702}          & 433.001          & \multicolumn{1}{c|}{22.824}          & \multicolumn{1}{c|}{0.749}          & \multicolumn{1}{c|}{0.805}          & 295.265         & \multicolumn{1}{c|}{28.202}          & \multicolumn{1}{c|}{0.890}          & \multicolumn{1}{c|}{0.908}          & 155.596         & 1446.937 \\ \hline
			PSTNN          & \multicolumn{1}{c|}{16.177}          & \multicolumn{1}{c|}{0.195}          & \multicolumn{1}{c|}{0.588}          & 607.694          & \multicolumn{1}{c|}{22.427}          & \multicolumn{1}{c|}{0.437}          & \multicolumn{1}{c|}{0.722}          & 308.484         & \multicolumn{1}{c|}{29.590}          & \multicolumn{1}{c|}{0.767}          & \multicolumn{1}{c|}{0.870}          & 137.052         & 304.501  \\ \hline
			FTNN           & \multicolumn{1}{c|}{24.881}          & \multicolumn{1}{c|}{0.694}          & \multicolumn{1}{c|}{0.836}          & 231.536          & \multicolumn{1}{c|}{28.324}          & \multicolumn{1}{c|}{0.826}          & \multicolumn{1}{c|}{0.895}          & 152.216         & \multicolumn{1}{c|}{32.690}          & \multicolumn{1}{c|}{0.923}          & \multicolumn{1}{c|}{0.945}          & 90.268          & 3510.595 \\ \hline
			WSTNN          & \multicolumn{1}{c|}{25.533}          & \multicolumn{1}{c|}{0.708}          & \multicolumn{1}{c|}{0.825}          & 211.248          & \multicolumn{1}{c|}{29.043}          & \multicolumn{1}{c|}{0.836}          & \multicolumn{1}{c|}{0.887}          & 139.577         & \multicolumn{1}{c|}{33.488}          & \multicolumn{1}{c|}{0.928}          & \multicolumn{1}{c|}{0.940}          & 82.945          & 724.544  \\ \hline
			NMCP           & \multicolumn{1}{c|}{28.874}          & \multicolumn{1}{c|}{0.814}          & \multicolumn{1}{c|}{0.871}          & 140.651          & \multicolumn{1}{c|}{32.338}          & \multicolumn{1}{c|}{0.902}          & \multicolumn{1}{c|}{0.919}          & 94.670          & \multicolumn{1}{c|}{35.822}          & \multicolumn{1}{c|}{0.952}          & \multicolumn{1}{c|}{0.954}          & 62.815          & 821.349  \\ \hline
			EMCP           & \multicolumn{1}{c|}{29.289}          & \multicolumn{1}{c|}{0.808}          & \multicolumn{1}{c|}{0.874}          & 132.805          & \multicolumn{1}{c|}{32.985}          & \multicolumn{1}{c|}{0.900}          & \multicolumn{1}{c|}{0.922}          & 86.655          & \multicolumn{1}{c|}{37.129}          & \multicolumn{1}{c|}{0.958}          & \multicolumn{1}{c|}{0.960}          & 53.454          & 1061.362 \\ \hline
			BEMCP & \multicolumn{1}{c|}{\textbf{29.734}} & \multicolumn{1}{c|}{\textbf{0.834}} & \multicolumn{1}{c|}{\textbf{0.883}} & \textbf{126.610} & \multicolumn{1}{c|}{\textbf{33.392}} & \multicolumn{1}{c|}{\textbf{0.912}} & \multicolumn{1}{c|}{\textbf{0.928}} & \textbf{82.986} & \multicolumn{1}{c|}{\textbf{37.194}} & \multicolumn{1}{c|}{\textbf{0.960}} & \multicolumn{1}{c|}{\textbf{0.961}} & \textbf{53.102} & 1108.386 \\ \hline
	\end{tabular}}\label{MRITC1}
\end{table*}
\subsection{CV completion}

\begin{figure*}[!h] %这里使用的是强制位置, 除非真的放不下, 不然就是写在哪里图就放在哪里, 不会乱动
	\centering  %图片全局居中
	\vspace{0cm} %设置与上面正文的距离
	%	\subfigtopskip=2pt %设置子图与上面正文或别的内容的距离
	%	\subfigbottomskip=2pt %设置第二行子图与第一行子图的距离, 即下面的头与上面的脚的距离
	%	\subfigcapskip=-5pt %设置子图与子标题之间的距离
	\subfloat[]{
		\begin{minipage}[b]{0.076\linewidth}
			\includegraphics[width=1\linewidth]{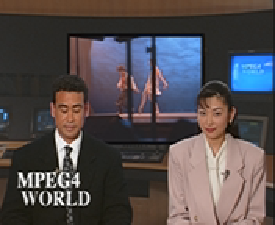}\\

			\includegraphics[width=1\linewidth]{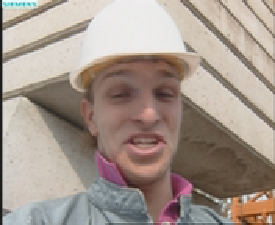}\\

			\includegraphics[width=1\linewidth]{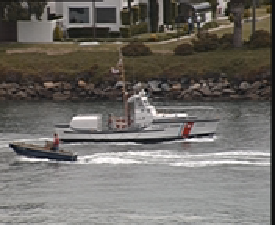}
	\end{minipage}}
	\subfloat[]{
		\begin{minipage}[b]{0.076\linewidth}
			\includegraphics[width=1\linewidth]{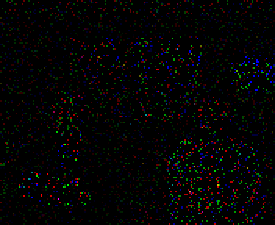}\\

			\includegraphics[width=1\linewidth]{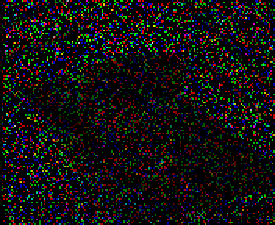}\\

			\includegraphics[width=1\linewidth]{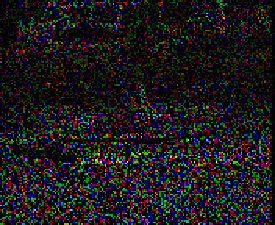}
	\end{minipage}}
	\subfloat[]{
		\begin{minipage}[b]{0.076\linewidth}
			\includegraphics[width=1\linewidth]{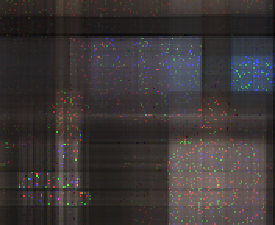}\\

			\includegraphics[width=1\linewidth]{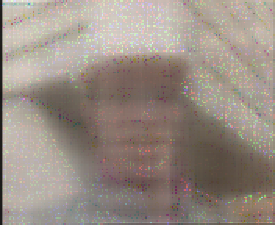}\\

			\includegraphics[width=1\linewidth]{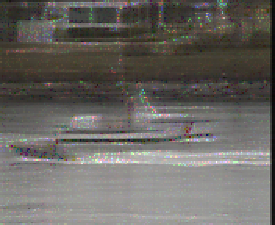}
	\end{minipage}}
	\subfloat[]{
		\begin{minipage}[b]{0.076\linewidth}
			\includegraphics[width=1\linewidth]{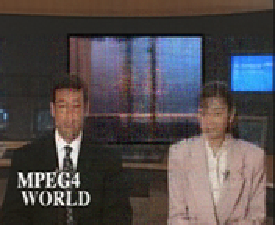}\\

			\includegraphics[width=1\linewidth]{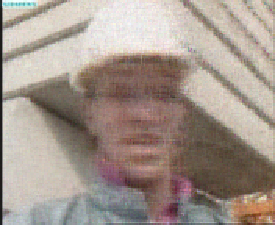}\\

			\includegraphics[width=1\linewidth]{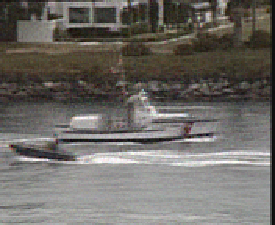}
	\end{minipage}}
	\subfloat[]{
		\begin{minipage}[b]{0.076\linewidth}
			\includegraphics[width=1\linewidth]{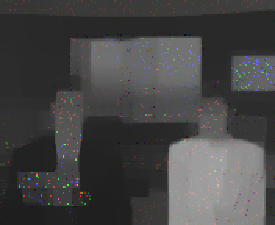}\\

			\includegraphics[width=1\linewidth]{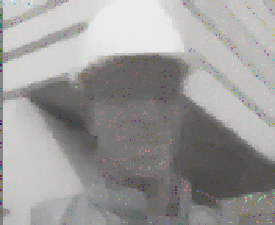}\\

			\includegraphics[width=1\linewidth]{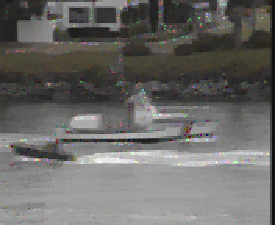}
	\end{minipage}}
	\subfloat[]{
		\begin{minipage}[b]{0.076\linewidth}
			\includegraphics[width=1\linewidth]{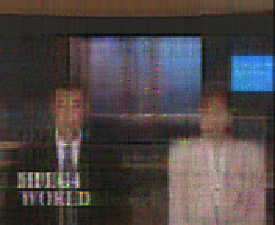}\\

			\includegraphics[width=1\linewidth]{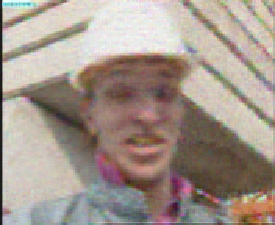}\\

			\includegraphics[width=1\linewidth]{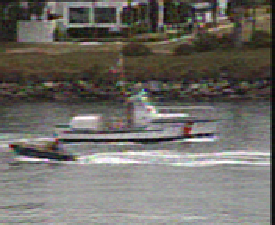}
	\end{minipage}}
	\subfloat[]{
		\begin{minipage}[b]{0.076\linewidth}
			\includegraphics[width=1\linewidth]{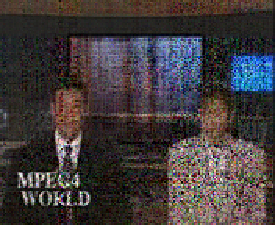}\\

			\includegraphics[width=1\linewidth]{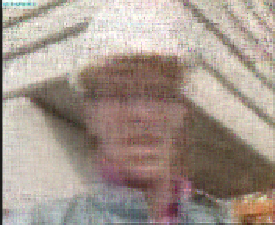}\\

			\includegraphics[width=1\linewidth]{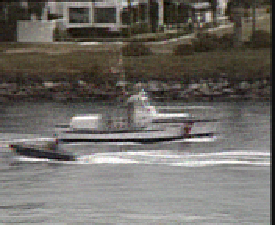}
	\end{minipage}}
	\subfloat[]{
		\begin{minipage}[b]{0.076\linewidth}
			\includegraphics[width=1\linewidth]{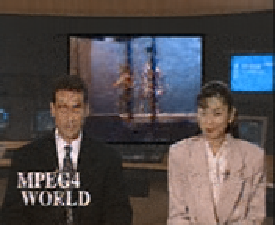}\\

			\includegraphics[width=1\linewidth]{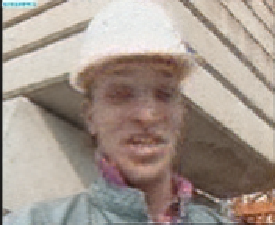}\\

			\includegraphics[width=1\linewidth]{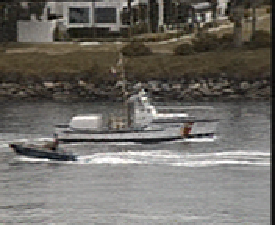}
	\end{minipage}}
	\subfloat[]{
		\begin{minipage}[b]{0.076\linewidth}
			\includegraphics[width=1\linewidth]{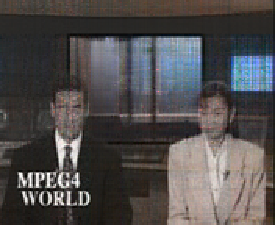}\\

			\includegraphics[width=1\linewidth]{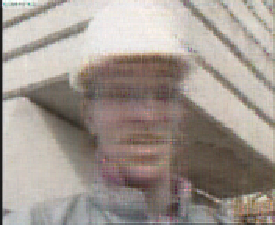}\\

			\includegraphics[width=1\linewidth]{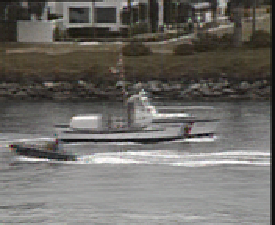}
	\end{minipage}}
	\subfloat[]{
		\begin{minipage}[b]{0.076\linewidth}
			\includegraphics[width=1\linewidth]{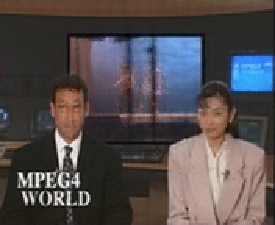}\\

			\includegraphics[width=1\linewidth]{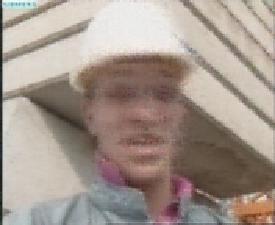}\\

			\includegraphics[width=1\linewidth]{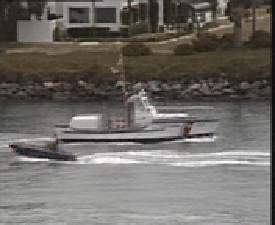}
	\end{minipage}}
	\subfloat[]{
		\begin{minipage}[b]{0.076\linewidth}
			\includegraphics[width=1\linewidth]{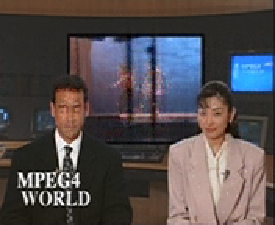}\\

			\includegraphics[width=1\linewidth]{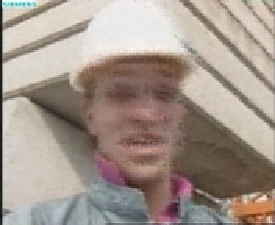}\\

			\includegraphics[width=1\linewidth]{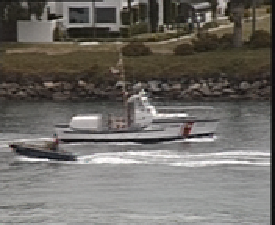}
	\end{minipage}}
	
	\caption{(a) Original image. (b) Observed image. (c) HaLRTC. (d) TNN. (e) LRTCTV-I. (f) PSTNN. (g) FTNN. (h) WSTNN. (i) NMCP. (j) EMCP. (k) BEMCP.  SR: top row is 5\%, middle row is 10\% and last row is 20\%. The rows of CVs are in order: the 10th frame of news, the 30th frame of foreman, and the 45th frame of coastguard.}
	\label{CVTC}
\end{figure*}

We test seven CVs\footnote{http://trace.eas.asu.edu/yuv/}(respectively named news, akiyo, foreman, hall, highway, container, coastguard) of size $144 \times 176 \times 3 \times 50$. Firstly, we demonstrate the visual results of 7 CVs in our experiment in Fig.\ref{CVTC}, in which the number of frames and sampling rate corresponding to each CV are described in the caption of Fig.\ref{CVTC}. It is not hard to see from the picture that the recovery of our method on the vision effect is more better. Furthermore, we list the average quantitative results of 7 CVs in Table \ref{CVTC1}. At this time, the suboptimal method is the EMCP. When the sampling rate is 5\%, the PSNR value of our method is 0.5dB higher than it. In addition, at the sampling rate of 5\% and 10\%, the PSNR value of all three methods is at least 3db higher than that of the WSTNN method. More experimental results are available in Appendix B.
\begin{table*}[]
	\caption{The average PSNR, SSIM, FSIM and ERGAS values for 7 CVs tested by observed and the nine utilized LRTC methods.}
	\resizebox{\textwidth}{!}{
		\begin{tabular}{|c|cccc|cccc|cccc|c|}
			\hline
			SR             & \multicolumn{4}{c|}{5\%}                                                                                                           & \multicolumn{4}{c|}{10\%}                                                                                                          & \multicolumn{4}{c|}{20\%}                                                                                                          & Time(s) \\ \hline
			Method         & \multicolumn{1}{c|}{PSNR}            & \multicolumn{1}{c|}{SSIM}           & \multicolumn{1}{c|}{FSIM}           & ERGAS           & \multicolumn{1}{c|}{PSNR}            & \multicolumn{1}{c|}{SSIM}           & \multicolumn{1}{c|}{FSIM}           & ERGAS           & \multicolumn{1}{c|}{PSNR}            & \multicolumn{1}{c|}{SSIM}           & \multicolumn{1}{c|}{FSIM}           & ERGAS           &         \\ \hline
			Observed       & \multicolumn{1}{c|}{5.793}           & \multicolumn{1}{c|}{0.011}          & \multicolumn{1}{c|}{0.420}          & 1194.940        & \multicolumn{1}{c|}{6.028}           & \multicolumn{1}{c|}{0.019}          & \multicolumn{1}{c|}{0.423}          & 1163.038        & \multicolumn{1}{c|}{6.540}           & \multicolumn{1}{c|}{0.034}          & \multicolumn{1}{c|}{0.429}          & 1096.477        & 0.000   \\ \hline
			HaLRTC         & \multicolumn{1}{c|}{17.336}          & \multicolumn{1}{c|}{0.488}          & \multicolumn{1}{c|}{0.695}          & 329.173         & \multicolumn{1}{c|}{21.141}          & \multicolumn{1}{c|}{0.622}          & \multicolumn{1}{c|}{0.774}          & 214.642         & \multicolumn{1}{c|}{24.981}          & \multicolumn{1}{c|}{0.772}          & \multicolumn{1}{c|}{0.862}          & 137.634         & 13.980  \\ \hline
			TNN            & \multicolumn{1}{c|}{27.033}          & \multicolumn{1}{c|}{0.771}          & \multicolumn{1}{c|}{0.886}          & 113.454         & \multicolumn{1}{c|}{30.453}          & \multicolumn{1}{c|}{0.855}          & \multicolumn{1}{c|}{0.928}          & 79.511          & \multicolumn{1}{c|}{33.697}          & \multicolumn{1}{c|}{0.910}          & \multicolumn{1}{c|}{0.955}          & 56.639          & 44.979  \\ \hline
			LRTCTV-I       & \multicolumn{1}{c|}{19.497}          & \multicolumn{1}{c|}{0.579}          & \multicolumn{1}{c|}{0.692}          & 272.767         & \multicolumn{1}{c|}{21.205}          & \multicolumn{1}{c|}{0.655}          & \multicolumn{1}{c|}{0.771}          & 228.491         & \multicolumn{1}{c|}{25.812}          & \multicolumn{1}{c|}{0.817}          & \multicolumn{1}{c|}{0.881}          & 126.740         & 281.989 \\ \hline
			PSTNN          & \multicolumn{1}{c|}{16.151}          & \multicolumn{1}{c|}{0.312}          & \multicolumn{1}{c|}{0.664}          & 364.981         & \multicolumn{1}{c|}{27.897}          & \multicolumn{1}{c|}{0.778}          & \multicolumn{1}{c|}{0.890}          & 102.835         & \multicolumn{1}{c|}{33.258}          & \multicolumn{1}{c|}{0.906}          & \multicolumn{1}{c|}{0.952}          & 58.772          & 44.823  \\ \hline
			FTNN           & \multicolumn{1}{c|}{25.286}          & \multicolumn{1}{c|}{0.766}          & \multicolumn{1}{c|}{0.871}          & 137.494         & \multicolumn{1}{c|}{28.544}          & \multicolumn{1}{c|}{0.858}          & \multicolumn{1}{c|}{0.917}          & 93.393          & \multicolumn{1}{c|}{32.214}          & \multicolumn{1}{c|}{0.924}          & \multicolumn{1}{c|}{0.954}          & 61.414          & 373.005 \\ \hline
			WSTNN          & \multicolumn{1}{c|}{29.257}          & \multicolumn{1}{c|}{0.872}          & \multicolumn{1}{c|}{0.920}          & 88.184          & \multicolumn{1}{c|}{32.635}          & \multicolumn{1}{c|}{0.924}          & \multicolumn{1}{c|}{0.952}          & 62.072          & \multicolumn{1}{c|}{36.557}          & \multicolumn{1}{c|}{0.960}          & \multicolumn{1}{c|}{0.975}          & 40.820          & 197.168 \\ \hline
			NMCP           & \multicolumn{1}{c|}{30.432}          & \multicolumn{1}{c|}{0.885}          & \multicolumn{1}{c|}{0.933}          & 77.007          & \multicolumn{1}{c|}{33.934}          & \multicolumn{1}{c|}{0.933}          & \multicolumn{1}{c|}{0.961}          & 52.983          & \multicolumn{1}{c|}{37.399}          & \multicolumn{1}{c|}{0.963}          & \multicolumn{1}{c|}{0.980}          & 35.450          & 211.281 \\ \hline
			EMCP           & \multicolumn{1}{c|}{30.768}          & \multicolumn{1}{c|}{0.887}          & \multicolumn{1}{c|}{0.937}          & 74.465          & \multicolumn{1}{c|}{34.512}          & \multicolumn{1}{c|}{0.934}          & \multicolumn{1}{c|}{0.964}          & 50.102          & \multicolumn{1}{c|}{38.204}          & \multicolumn{1}{c|}{0.964}          & \multicolumn{1}{c|}{0.982}          & 32.499          & 226.812 \\ \hline
			BEMCP & \multicolumn{1}{c|}{\textbf{31.284}} & \multicolumn{1}{c|}{\textbf{0.893}} & \multicolumn{1}{c|}{\textbf{0.942}} & \textbf{70.978} & \multicolumn{1}{c|}{\textbf{34.683}} & \multicolumn{1}{c|}{\textbf{0.935}} & \multicolumn{1}{c|}{\textbf{0.965}} & \textbf{49.164} & \multicolumn{1}{c|}{\textbf{38.244}} & \multicolumn{1}{c|}{\textbf{0.964}} & \multicolumn{1}{c|}{\textbf{0.982}} & \textbf{32.285} & 247.195 \\ \hline
	\end{tabular}}\label{CVTC1}
\end{table*}
\subsection{Discussions}
\subsubsection{Model Analysis}
The two parameters $\alpha$ and $\rho$ are included in our proposed BEMCP model. In addition, the initial values of the variables $\upsilon$, $\Lambda$, and $\Gamma$ also strongly influence on the efficient solution of the BEMCP model. The parameter values of $\alpha$ and $\rho$ follow the settings in the N-tubal rank \cite{2020170}. The initial value of the variable $\upsilon_{0}$ is $\upsilon_{0}=\Lambda_{0}\star\Gamma_{0}$. The specific data affects the initial value design of $\Lambda$ and $\Gamma$. The optimal initial value design can be found in Appendix C.
\subsubsection{Convergency Behaviours}
\begin{figure*}
	\centering
	\subfloat[MSI]{
		%		\label{level.sub.2}
		\includegraphics[width=0.3\linewidth]{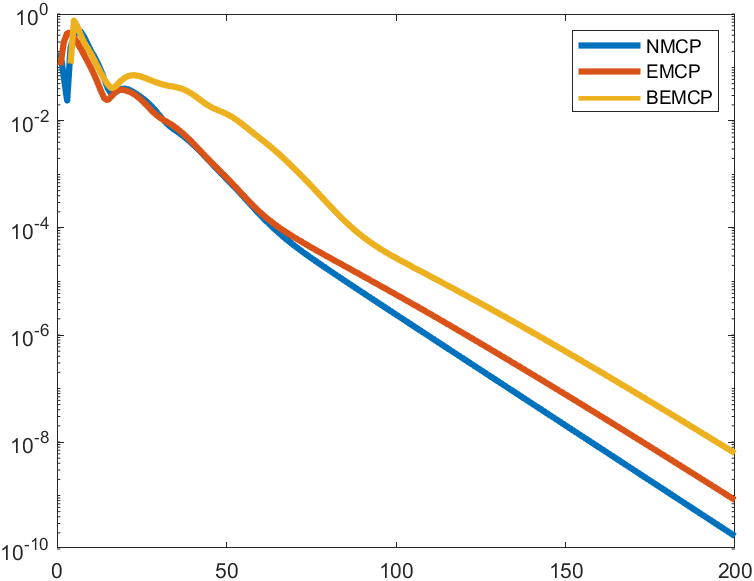}
	}
	\subfloat[MRI]{
	%		\label{level.sub.2}
	\includegraphics[width=0.3\linewidth]{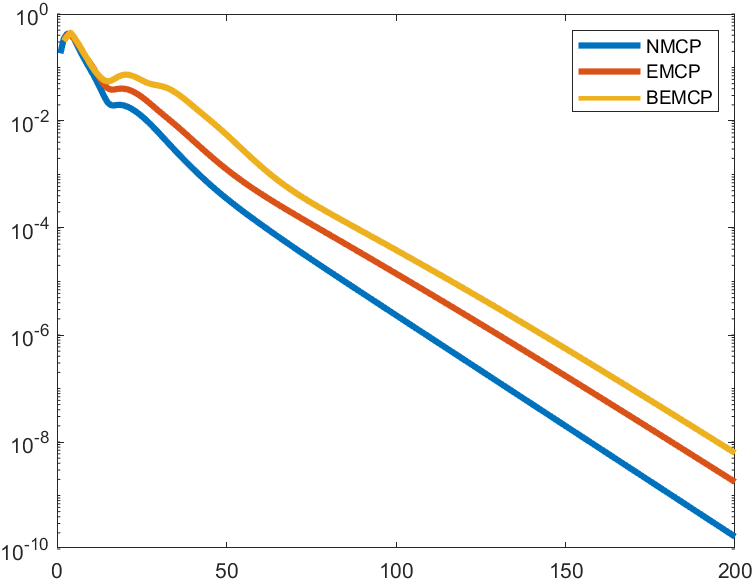}
}
	\subfloat[CV]{
	%		\label{level.sub.2}
	\includegraphics[width=0.3\linewidth]{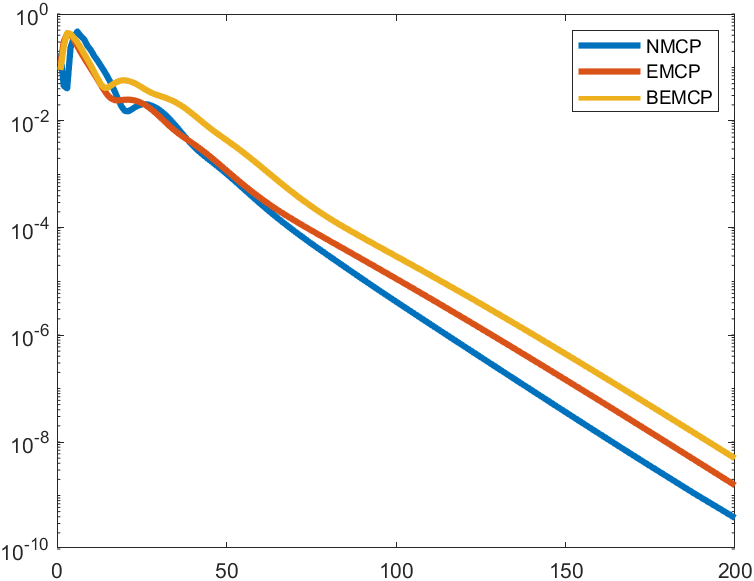}
}
	\caption{The convergence behaviours of LRTC algorithm, with respect to MSI, MRI, CV data.}
	\label{convergence}
\end{figure*}

We take the LRTC of MSI, MRI, CV data as examples to illustrate the convergence behavior of the three algorithms under 5\% sampling rate. We have drawn $\|\mathcal{X}^{k+1}-\mathcal{X}^{k}\|_{\infty}$ for each iteration in Fig.\ref{convergence}. It can be seen that our algorithm converges stably and quickly.

\section{CONCLUSION}
This paper proposes a new structural equivalence theorem called the BEMCP theorem, which converts two constant parameters in the MCP function into variables so that it can be adaptively updated with the iterative update of tensor singular values in the algorithm. Compared with other existing methods, the effect is remarkable, and it is slightly better than the same type of the NMCP and EMCP methods. On this basis, we give the BEMCP model for solving the LRTC problem. Extensive experiments show that our method can achieve better visual and numerical quantitative results than the comparison methods.

% or


%\appendices

%\section{}

%\section*{Acknowledgment}
%\cite{}

%The authors would like to thank

\ifCLASSOPTIONcaptionsoff
  \newpage
\fi

\bibliography{bibtex/bare_jrnl_cs1}

\begin{thebibliography}{10}
\providecommand{\url}[1]{#1}
\csname url@samestyle\endcsname
\providecommand{\newblock}{\relax}
\providecommand{\bibinfo}[2]{#2}
\providecommand{\BIBentrySTDinterwordspacing}{\spaceskip=0pt\relax}
\providecommand{\BIBentryALTinterwordstretchfactor}{4}
\providecommand{\BIBentryALTinterwordspacing}{\spaceskip=\fontdimen2\font plus
\BIBentryALTinterwordstretchfactor\fontdimen3\font minus
  \fontdimen4\font\relax}
\providecommand{\BIBforeignlanguage}[2]{{%
\expandafter\ifx\csname l@#1\endcsname\relax
\typeout{** WARNING: IEEEtran.bst: No hyphenation pattern has been}%
\typeout{** loaded for the language `#1'. Using the pattern for}%
\typeout{** the default language instead.}%
\else
\language=\csname l@#1\endcsname
\fi
#2}}
\providecommand{\BIBdecl}{\relax}
\BIBdecl

\bibitem{1472018163}
Y.-M. Huang, H.-Y. Yan, Y.-W. Wen, and X.~Yang, ``{R}ank minimization with
  applications to image noise removal,'' \emph{Information Sciences}, vol. 429,
  pp. 147--163, 2018.

\bibitem{3762018t397}
B.~Madathil and S.~N. George, ``{T}wist tensor total variation
  regularized-reweighted nuclear norm based tensor completion for video missing
  area recovery,'' \emph{Information Sciences}, vol. 423, pp. 376--397, 2018.

\bibitem{7562018767}
Y.~Wang, D.~Meng, and M.~Yuan, ``{S}parse recovery: from vectors to tensors,''
  \emph{National Science Review}, vol.~5, no.~5, pp. 756--767, 2018.

\bibitem{1372020149}
X.-L. Zhao, W.-H. Xu, T.-X. Jiang, Y.~Wang, and M.~K. Ng, ``{D}eep
  plug-and-play prior for low-rank tensor completion,'' \emph{Neurocomputing},
  vol. 400, pp. 137--149, 2020.

\bibitem{8359412}
S.~Li, R.~Dian, L.~Fang, and J.~M. Bioucas-Dias, ``{F}using {H}yperspectral and
  {M}ultispectral {I}mages via {C}oupled {S}parse {T}ensor {F}actorization,''
  \emph{IEEE Transactions on Image Processing}, vol.~27, no.~8, pp. 4118--4130,
  2018.

\bibitem{7467446}
X.~Fu, W.-K. Ma, J.~M. Bioucas-Dias, and T.-H. Chan, ``{S}emiblind
  {H}yperspectral {U}nmixing in the {P}resence of {S}pectral {L}ibrary
  {M}ismatches,'' \emph{IEEE Transactions on Geoscience and Remote Sensing},
  vol.~54, no.~9, pp. 5171--5184, 2016.

\bibitem{8657368}
J.~Xue, Y.~Zhao, W.~Liao, and J.~C.-W. Chan, ``{N}onlocal {L}ow-{R}ank
  {R}egularized {T}ensor {D}ecomposition for {H}yperspectral {I}mage
  {D}enoising,'' \emph{IEEE Transactions on Geoscience and Remote Sensing},
  vol.~57, no.~7, pp. 5174--5189, 2019.

\bibitem{8941238}
J.~Xue, Y.~Zhao, W.~Liao, J.~C.-W. Chan, and S.~G. Kong, ``{E}nhanced
  {S}parsity {P}rior {M}odel for {L}ow-{R}ank {T}ensor {C}ompletion,''
  \emph{IEEE Transactions on Neural Networks and Learning Systems}, vol.~31,
  no.~11, pp. 4567--4581, 2020.

\bibitem{1242020783}
J.-H. Yang, X.-L. Zhao, T.-Y. Ji, T.-H. Ma, and T.-Z. Huang, ``{L}ow-rank
  tensor train for tensor robust principal component analysis,'' \emph{Applied
  Mathematics and Computation}, vol. 367, p. 124783, 2020.

\bibitem{4032018417}
T.-X. Jiang, T.-Z. Huang, X.-L. Zhao, T.-Y. Ji, and L.-J. Deng, ``{M}atrix
  factorization for low-rank tensor completion using framelet prior,''
  \emph{Information Sciences}, vol. 436, pp. 403--417, 2018.

\bibitem{9412019964}
M.~Ding, T.-Z. Huang, T.-Y. Ji, X.-L. Zhao, and J.-H. Yang, ``{L}ow-rank tensor
  completion using matrix factorization based on tensor train rank and total
  variation,'' \emph{Journal of Scientific Computing}, vol.~81, no.~2, pp.
  941--964, 2019.

\bibitem{5032019109}
J.~Xue, Y.~Zhao, W.~Liao, and J.~{Cheung-Wai Chan}, ``{N}onconvex tensor rank
  minimization and its applications to tensor recovery,'' \emph{Information
  Sciences}, vol. 503, pp. 109--128, 2019.

\bibitem{8319458}
I.~Kajo, N.~Kamel, Y.~Ruichek, and A.~S. Malik, ``{SVD}-{B}ased
  {T}ensor-{C}ompletion {T}echnique for {B}ackground {I}nitialization,''
  \emph{IEEE Transactions on Image Processing}, vol.~27, no.~6, pp. 3114--3126,
  2018.

\bibitem{7488247}
W.~Cao, Y.~Wang, J.~Sun, D.~Meng, C.~Yang, A.~Cichocki, and Z.~Xu, ``{T}otal
  {V}ariation {R}egularized {T}ensor {RPCA} for {B}ackground {S}ubtraction
  {F}rom {C}ompressive {M}easurements,'' \emph{IEEE Transactions on Image
  Processing}, vol.~25, no.~9, pp. 4075--4090, 2016.

\bibitem{8237537}
W.~Wei, L.~Yi, Q.~Xie, Q.~Zhao, D.~Meng, and Z.~Xu, ``{S}hould {W}e {E}ncode
  {R}ain {S}treaks in {V}ideo as {D}eterministic or {S}tochastic?'' in
  \emph{2017 IEEE International Conference on Computer Vision (ICCV)}, 2017,
  pp. 2535--2544.

\bibitem{8578793}
M.~Li, Q.~Xie, Q.~Zhao, W.~Wei, S.~Gu, J.~Tao, and D.~Meng, ``{V}ideo {R}ain
  {S}treak {R}emoval by {M}ultiscale {C}onvolutional {S}parse {C}oding,'' in
  \emph{2018 IEEE/CVF Conference on Computer Vision and Pattern Recognition},
  2018, pp. 6644--6653.

\bibitem{7010937}
Q.~Zhao, L.~Zhang, and A.~Cichocki, ``{B}ayesian {CP} {F}actorization of
  {I}ncomplete {T}ensors with {A}utomatic {R}ank {D}etermination,'' \emph{IEEE
  Transactions on Pattern Analysis and Machine Intelligence}, vol.~37, no.~9,
  pp. 1751--1763, 2015.

\bibitem{7676397}
T.~Yokota, N.~Lee, and A.~Cichocki, ``{R}obust {M}ultilinear {T}ensor {R}ank
  {E}stimation {U}sing {H}igher {O}rder {S}ingular {V}alue {D}ecomposition and
  {I}nformation {C}riteria,'' \emph{IEEE Transactions on Signal Processing},
  vol.~65, no.~5, pp. 1196--1206, 2017.

\bibitem{41201156}
E.~Acar, D.~M. Dunlavy, T.~G. Kolda, and M.~M{\o}rup, ``{S}calable tensor
  factorizations for incomplete data,'' \emph{Chemometrics and Intelligent
  Laboratory Systems}, vol. 106, no.~1, pp. 41--56, 2011.

\bibitem{36232017}
P.~Tichavsk{\`y}, A.-H. Phan, and A.~Cichocki, ``{N}umerical {CP} decomposition
  of some difficult tensors,'' \emph{Journal of Computational and Applied
  Mathematics}, vol. 317, pp. 362--370, 2017.

\bibitem{64201881}
Y.-F. Li, K.~Shang, and Z.-H. Huang, ``{L}ow {T}ucker rank tensor recovery via
  {ADMM} based on exact and inexact iteratively reweighted algorithms,''
  \emph{Journal of Computational and Applied Mathematics}, vol. 331, pp.
  64--81, 2018.

\bibitem{7460200}
X.~Li, M.~K. Ng, G.~Cong, Y.~Ye, and Q.~Wu, ``{MR}-{NTD}: {M}anifold
  {R}egularization {N}onnegative {T}ucker {D}ecomposition for {T}ensor {D}ata
  {D}imension {R}eduction and {R}epresentation,'' \emph{IEEE Transactions on
  Neural Networks and Learning Systems}, vol.~28, no.~8, pp. 1787--1800, 2017.

\bibitem{6909886}
Z.~Zhang, G.~Ely, S.~Aeron, N.~Hao, and M.~Kilmer, ``{N}ovel {M}ethods for
  {M}ultilinear {D}ata {C}ompletion and {D}e-noising {B}ased on
  {T}ensor-{SVD},'' \emph{2014 IEEE Conference on Computer Vision and Pattern
  Recognition}, pp. 3842--3849, 2014.

\bibitem{139201360}
C.~J. Hillar and L.-H. Lim, ``{M}ost tensor problems are {NP}-hard,''
  \emph{Journal of the ACM (JACM)}, vol.~60, no.~6, pp. 1--39, 2013.

\bibitem{2020170}
Y.-B. Zheng, T.-Z. Huang, X.-L. Zhao, T.-X. Jiang, T.-Y. Ji, and T.-H. Ma,
  ``{T}ensor {N}-tubal rank and its convex relaxation for low-rank tensor
  recovery,'' \emph{Information Sciences}, vol. 532, pp. 170--189, 2020.

\bibitem{448120174494}
I.~Selesnick, ``{S}parse regularization via convex analysis,'' \emph{IEEE
  Transactions on Signal Processing}, vol.~65, no.~17, pp. 4481--4494, 2017.

\bibitem{2271995234}
B.~K. Natarajan, ``{S}parse approximate solutions to linear systems,''
  \emph{SIAM Journal on Computing}, vol.~24, no.~2, pp. 227--234, 1995.

\bibitem{211720122130}
Y.~Hu, D.~Zhang, J.~Ye, X.~Li, and X.~He, ``{F}ast and {A}ccurate {M}atrix
  {C}ompletion via {T}runcated {N}uclear {N}orm {R}egularization,'' \emph{IEEE
  Transactions on Pattern Analysis and Machine Intelligence}, vol.~35, no.~9,
  pp. 2117--2130, 2013.

\bibitem{4712010501}
B.~Recht, M.~Fazel, and P.~A. Parrilo, ``{G}uaranteed minimum-rank solutions of
  linear matrix equations via nuclear norm minimization,'' \emph{SIAM Review},
  vol.~52, no.~3, pp. 471--501, 2010.

\bibitem{qiu2021nonlocal}
D.~Qiu, M.~Bai, M.~K. Ng, and X.~Zhang, ``{N}onlocal robust tensor recovery
  with nonconvex regularization,'' \emph{Inverse Problems}, vol.~37, no.~3, p.
  035001, 2021.

\bibitem{2920151}
C.~Lu, C.~Zhu, C.~Xu, S.~Yan, and Z.~Lin, ``Generalized singular value
  thresholding,'' \emph{Proceedings of the AAAI Conference on Artificial
  Intelligence}, vol.~29, no.~1, 2015.

\bibitem{8942010942}
C.-H. Zhang, ``Nearly unbiased variable selection under minimax concave
  penalty,'' \emph{The Annals of statistics}, vol.~38, no.~2, pp. 894--942,
  2010.

\bibitem{2132021245}
P.~K. Pokala, R.~V. Hemadri, and C.~S. Seelamantula, ``{I}teratively
  {R}eweighted {M}inimax-{C}oncave {P}enalty {M}inimization for {A}ccurate
  {L}ow-rank {P}lus {S}parse {M}atrix {D}ecomposition,'' \emph{IEEE
  Transactions on Pattern Analysis and Machine Intelligence}, 2021.

\bibitem{683201156}
S.~Boyd, N.~Parikh, and E.~Chu, \emph{{D}istributed optimization and
  statistical learning via the alternating direction method of
  multipliers}.\hskip 1em plus 0.5em minus 0.4em\relax Now Publishers Inc,
  2011.

\bibitem{6122011620}
Z.~Lin, R.~Liu, and Z.~Su, ``{L}inearized {A}lternating {D}irection {M}ethod
  with {A}daptive {P}enalty for {L}ow-{R}ank {R}epresentation,'' \emph{Advances
  in Neural Information Processing Systems}, vol.~24, pp. 612--620, 2011.

\bibitem{6416568}
M.~E. Kilmer and C.~D. Martin, ``{F}actorization strategies for third-order
  tensors,'' \emph{Linear Algebra and its Applications}, vol. 435, no.~3, pp.
  641--658, 2011.

\bibitem{8606166}
C.~Lu, J.~Feng, Y.~Chen, W.~Liu, Z.~Lin, and S.~Yan, ``{T}ensor {R}obust
  {P}rincipal {C}omponent {A}nalysis with a {N}ew {T}ensor {N}uclear {N}orm,''
  \emph{IEEEsactions on Pattern Analysis and Machine Intelligence}, vol.~42,
  no.~4, pp. 925--938, 2020.

\bibitem{mirsky1975trace}
L.~Mirsky, ``{A} trace inequality of {J}ohn von {N}eumann,'' \emph{Monatshefte
  f{\"u}r mathematik}, vol.~79, no.~4, pp. 303--306, 1975.

\bibitem{lin2010augmented}
Z.~Lin, M.~Chen, and Y.~Ma, ``{T}he augmented lagrange multiplier method for
  exact recovery of corrupted low-rank matrices,'' \emph{arXiv preprint
  arXiv:1009.5055}, 2010.

\bibitem{1284395}
Z.~Wang, A.~Bovik, H.~Sheikh, and E.~Simoncelli, ``{I}mage quality assessment:
  from error visibility to structural similarity,'' \emph{IEEE Transactions on
  Image Processing}, vol.~13, no.~4, pp. 600--612, 2004.

\bibitem{5705575}
L.~Zhang, L.~Zhang, X.~Mou, and D.~Zhang, ``{FSIM}: {A} {F}eature {S}imilarity
  {I}ndex for {I}mage {Q}uality {A}ssessment,'' \emph{IEEE Transactions on
  Image Processing}, vol.~20, no.~8, pp. 2378--2386, 2011.

\bibitem{2432002352}
L.~Wald, \emph{{D}ata fusion: definitions and architectures: fusion of images
  of different spatial resolutions}.\hskip 1em plus 0.5em minus 0.4em\relax
  Presses des MINES, 2002.

\bibitem{6138863}
J.~Liu, P.~Musialski, P.~Wonka, and J.~Ye, ``{T}ensor {C}ompletion for
  {E}stimating {M}issing {V}alues in {V}isual {D}ata,'' \emph{IEEE Transactions
  on Pattern Analysis and Machine Intelligence}, vol.~35, no.~1, pp. 208--220,
  2013.

\bibitem{3120171}
X.~Li, Y.~Ye, and X.~Xu, ``Low-rank tensor completion with total variation for
  visual data inpainting,'' \emph{Proceedings of the AAAI Conference on
  Artificial Intelligence}, vol.~31, no.~1, 2017.

\bibitem{7782758}
Z.~Zhang and S.~Aeron, ``{E}xact {T}ensor {C}ompletion {U}sing t-{SVD},''
  \emph{IEEE Transactions on Signal Processing}, vol.~65, no.~6, pp.
  1511--1526, 2017.

\bibitem{1122020112680}
T.-X. Jiang, T.-Z. Huang, X.-L. Zhao, and L.-J. Deng, ``{M}ulti-dimensional
  imaging data recovery via minimizing the partial sum of tubal nuclear norm,''
  \emph{Journal of Computational and Applied Mathematics}, vol. 372, p. 112680,
  2020.

\bibitem{9115254}
T.-X. Jiang, M.~K. Ng, X.-L. Zhao, and T.-Z. Huang, ``{F}ramelet
  {R}epresentation of {T}ensor {N}uclear {N}orm for {T}hird-{O}rder {T}ensor
  {C}ompletion,'' \emph{IEEE Transactions on Image Processing}, vol.~29, pp.
  7233--7244, 2020.

\end{thebibliography}
%\bibliography{bibtex/IEEEexample}
%\end{IEEEbiography}


%\newpage


%\vfill


\end{document}